%% file: main-3317-Kreutzer-finalversion.tex
\documentclass[11pt,a4paper]{article}
\usepackage{times,latexsym}
\usepackage{url}
\usepackage[T1]{fontenc}

\usepackage[acceptedWithA]{tacl2018v2}

\usepackage{arydshln}
\usepackage{booktabs}
\usepackage{rotating}
\usepackage{array}
\usepackage{hhline}

\usepackage{graphicx}
\usepackage{multirow}
\usepackage{subcaption}
\usepackage{amssymb,graphicx,stackengine,xcolor, amsmath}
\def\ucr{\scalebox{1}{\stackinset{c}{}{c}{-.1pt}{%
  \textcolor{white}{\sffamily\bfseries\small ?}}{%
  \rotatebox{45}{$\blacksquare$}}}}

\usepackage{xspace,mfirstuc,tabulary}

\newif\iftaclinstructions
\taclinstructionsfalse % AUTHORS: do NOT set this to true
\iftaclinstructions
\renewcommand{\confidential}{}
\renewcommand{\anonsubtext}{(No author info supplied here, for consistency with
TACL-submission anonymization requirements)}
\newcommand{\instr}
\fi

\iftaclpubformat % this "if" is set by the choice of options

\else

\fi

\title{Quality at a Glance:\\ An Audit of Web-Crawled Multilingual Datasets}

\author{ \bf
Julia Kreutzer$^{a,b}$, 
Isaac Caswell$^a$,  
Lisa Wang$^a$,   
Ahsan Wahab$^c$, 
Daan van Esch$^a$,\\
\bf 
Nasanbayar Ulzii-Orshikh$^{d}$, 
Allahsera Tapo$^{b,e}$, 
Nishant Subramani$^{b,\delta}$, 
Artem Sokolov$^a$, \\
\bf
Claytone Sikasote$^{b,g}$, 
Monang Setyawan$^h$,
Supheakmungkol Sarin$^{h}$,
\\
\bf 
Sokhar Samb$^{b,i}$,  
Benoît Sagot$^{j}$,
Clara Rivera$^a$, 
Annette Rios$^{k}$,  
Isabel Papadimitriou$^{l}$,  \\
\bf 
Salomey Osei$^{b,m}$, 
Pedro Ortiz Suarez$^{j, n}$, %Pedro Javier Ortiz Suárez$^{j, n}$, 
Iroro Orife$^{b,o}$, 
Kelechi Ogueji$^{b,p}$,  \\
\bf 
Andre Niyongabo Rubungo$^{b,q}$, 
Toan Q. Nguyen$^{r}$, 
Mathias Müller$^{k}$, 
André Müller$^{k}$,\\
\bf
Shamsuddeen Hassan Muhammad$^{b,s}$,  
Nanda Muhammad$^{h}$,
Ayanda Mnyakeni$^{h}$, \\
\bf
Jamshidbek Mirzakhalov$^{c,t}$,  
Tapiwanashe Matangira$^h$, 
Colin Leong$^{b}$,
Nze Lawson$^h$,  \\
\bf  
Sneha Kudugunta$^a$, 
Yacine Jernite$^{b,u}$, 
Mathias Jenny$^{k}$, 
Orhan Firat$^{a,c}$, 
 \\
\bf 
Bonaventure F. P. Dossou$^{b,v}$, 
Sakhile Dlamini$^h$, 
Nisansa de Silva$^{w}$,  
Sakine Çabuk Ballı$^{k}$, \\
\bf 
Stella Biderman$^{x}$, 
Alessia Battisti$^{k}$, 
Ahmed Baruwa$^{b,y}$, 
Ankur Bapna$^a$,  \\
\bf 
Pallavi Baljekar$^a$, 
Israel Abebe Azime$^{b,i}$, 
Ayodele Awokoya$^{b,z}$, 
Duygu Ataman$^{c, k}$,  \\
\bf 
Orevaoghene Ahia$^{b,\alpha}$, 
Oghenefego Ahia$^{h}$, 
Sweta Agrawal$^{\beta}$, 
Mofetoluwa Adeyemi$^{b,\gamma}$, \\ \\

$^a$Google Research, $^b$Masakhane NLP, $^c$Turkic Interlingua, $^d$Haverford College, \\
$^e$RobotsMali, $^f$Intel Labs, $^g$University of Zambia, $^h$Google,  $^i$AIMS-AMMI,\\
$^j$Inria,
 $^k$University of Zurich, $^l$Stanford University,\\
 $^m$Kwame Nkrumah University of Science and Technology,\\
 $^n$Sorbonne Université, $^o$Niger-Volta LTI, $^p$University of Waterloo\\
$^q$University of Electronic Science and Technology of China, $^r$University of Notre Dame, \\
$^{s}$Bayero University Kano, $^{t}$University of South Florida, $^{u}$Hugging Face,  \\
$^{v}$Jacobs University Bremen,
 $^{w}$University of Moratuwa, $^{x}$EleutherAI, \\
 $^{y}$Obafemi Awolowo University, $^{z}$University of Ibadan,
 $^{\alpha}$Instadeep, \\
$^{\beta}$University of Maryland,
$^{\gamma}$Defence Space Administration Abuja,\\
$^{\delta}$Allen Institute for Artificial Intelligence

}

\date{}

\begin{document}
\maketitle
\begin{abstract}
With the success of large-scale pre-training and multilingual modeling in Natural Language Processing (NLP), recent years have seen a proliferation of large, web-mined text datasets covering hundreds of languages. %However, to date there has been no systematic analysis of the quality of these publicly available datasets, or whether the datasets actually contain content in the languages they claim to represent. 
%In this work, 
We manually audit the quality of 205 language-specific corpora released with five major public datasets (CCAligned, ParaCrawl, WikiMatrix, OSCAR, mC4). %, and audit the correctness of language codes in a sixth (JW300). 
%We find that 
Lower-resource corpora have systematic issues: At least 15 corpora have no usable text, and a significant fraction contains less than 50\% sentences of acceptable quality. 
%Similarly, we find 
%An additional 
In addition, 
%83 corpora 
many are mislabeled or use nonstandard/ambiguous language codes. We demonstrate that these issues are easy to detect even for non-proficient speakers, 
%speakers of the languages in question, 
and supplement the human audit with automatic analyses.
%human judgements with automatic analyses.
%Inspired by our analysis, 
Finally, we recommend techniques to evaluate and improve multilingual corpora and discuss potential risks that come with low-quality data releases.
\end{abstract}

\section{Introduction}
Access to multilingual datasets for NLP research has vastly improved over the past years. A variety of web-derived collections for hundreds of languages is available for anyone to download, such as ParaCrawl~\citep{espla-etal-2019-paracrawl, banon-etal-2020-paracrawl}, WikiMatrix~\citep{schwenk-etal-2021-wikimatrix} CCAligned~\citep{el-kishky-etal-2020-ccaligned}, \mbox{OSCAR}~\citep{ortiz-suarez-etal-2019-asynchronous, ortiz-suarez-etal-2020-monolingual}, and several others. 
These have in turn enabled a variety of highly multilingual models, like mT5~\citep{xue-etal-2021-mt5}, M2M\nobreakdash-100~\citep{fan2020englishcentric}, M4~\citep{arivazhagan2019massively}.
% Recent approaches that employ large-scale multilingual systems have shown potential for some generalization across typologically different languages \cite{pires-etal-2019-multilingual,papadimitriou-etal-2021-deep}. .

Curating such datasets relies on the websites giving clues about the language of their contents (e.g. a language identifier in the URL) and on automatic language classification (LangID).
%and filtering tools, and a range of evaluation sets and metrics to compile high-quality datasets. 
It is commonly known that these automatically crawled and filtered datasets tend to have overall lower quality than hand-curated collections~\citep{koehn-etal-2020-findings}, but their quality is rarely measured directly, and is rather judged through the improvements they bring to downstream applications~\citep{schwenk-etal-2021-wikimatrix}. 

Building NLP technologies with automatically crawled datasets is promising. This is especially true for low-resource languages, because data scarcity is one of the major bottlenecks for deep learning approaches.
However, there is a problem: There exists very little research on evaluating both data collections and automatic crawling and filtering tools for low-resource languages.
As a result, although many low-resource languages are covered by the latest multilingual crawl data releases, their quality and thus usability is unknown. 

To shed light on the quality of data crawls for the lowest resource languages, we perform a manual data audit for 230 per-language subsets of five major crawled multilingual datasets:\footnote{Annotations are available for 
\href{https://storage.googleapis.com/huggingface-nlp/datasets/masakhane_audit_annotations/masakhane_language_audit.zip}{download} (last accessed: 12 Oct 2021).} 
CCAligned~\citep{el-kishky-etal-2020-ccaligned}, ParaCrawl~\citep{espla-etal-2019-paracrawl,banon-etal-2020-paracrawl}, WikiMatrix~\citep{schwenk-etal-2021-wikimatrix}, OSCAR~\citep{ortiz-suarez-etal-2019-asynchronous, ortiz-suarez-etal-2020-monolingual} and mC4~\citep{xue-etal-2021-mt5}. We propose solutions for effective, low-effort data auditing (Section~\ref{sec:audit}), including an error taxonomy. Our quantitative analysis reveals surprisingly low amounts of valid in-language data, and identifies systematic issues across datasets and languages. In addition, we find that a large number of datasets is labeled with nontransparent or incorrect language codes (Section~\ref{sec:codes}). This leads us to reflect on the potential harm of low-quality data releases for low-resource languages (Section~\ref{sec:risk}), and provide a set of recommendations for future multilingual data releases (Section~\ref{sec:recommendation}).

\begin{table*}[th!]
    \centering
    \resizebox{\textwidth}{!}{%
    \begin{tabular}{lccccc}
    \toprule
    & \multicolumn{3}{c}{\textbf{Parallel}} & \multicolumn{2}{c}{\textbf{Monolingual}} \\
    \cmidrule(lr){2-4} \cmidrule(lr){5-6}
    & \textbf{CCAligned} & \textbf{ParaCrawl v7.1} & \textbf{WikiMatrix} & \textbf{OSCAR} & \textbf{mC4}\\
    \midrule
     \#languages & 137 & 41 & 85 & 166 & 101 \\
     Source & CC 2013--2020 & selected websites & Wikipedia & CC 11/2018 & CC all\\ 
     Filtering level & document & sentence & sentence & document & document \\
     Langid & FastText & CLD2 & FastText & FastText & CLD3 \\
     Alignment & LASER & Vec/Hun/BLEU-Align & LASER & - & - \\
     Evaluation & TED-6 & WMT-5 & TED-45 & POS/DEP-5 & XTREME \\
    \bottomrule
    \end{tabular}%
    }
    \caption{Comparison of parallel and monolingual corpora extracted from web documents, including their downstream evaluation tasks. All parallel corpora are evaluated for machine translation (BLEU). TED-6: \texttt{da}, \texttt{cr}, \texttt{sl}, \texttt{sk}, \texttt{lt}, \texttt{et}; TED-45: 45-language subset of ~\citep{qi-etal-2018-pre}; WMT-5: \texttt{cs}, \texttt{de}, \texttt{fi}, \texttt{lv}, \texttt{ro}. POS/DEP-5: part-of-speech labeling and dependency parsing for \texttt{bg}, \texttt{ca}, \texttt{da}, \texttt{fi}, \texttt{id}.}
    \label{tab:corpora}
\end{table*}
% CC: CommonCrawl; TED-6: da, cr, sl, sk, lt, et; TED-50: TODO cite Qi et al. 2018; WMT-5: cs, de, fi, lv, ro.
%fastText: 176 languages, Wikipedia & Tatoeba

\section{Related Work}\label{sec:related}

Corpora collected by web crawlers are known to be noisy~\citep{junczys-dowmunt-2019-microsoft,luccioni-box}. In highly multilingual settings, past work found that web-crawls of lower-resource languages have serious issues, especially with segment-level LangID~\citep{caswell-etal-2020-language}. 
%Repeated studies have shown that 
Cleaning and filtering web-crawls can boost general language modeling~\citep{gao2020pile,GPT3,raffel2020exploring} and downstream task performance~\citep{moore-lewis-2010-intelligent,rarrick2011mt,xu-koehn-2017-zipporah,khayrallah-koehn-2018-impact,GPT3}.

As the scale of ML research grows, it becomes increasingly difficult to validate automatically collected and curated datasets~\citep{biderman2020pitfalls,prabhu2020large,bender2021dangers}.
%Data Quality Considerations for Big Data and Machine Learning: Going Beyond Data Cleaning and Transformations \citep{gudivada2017data}
Several works have focused on advancing methodologies and best practices to address these challenges. \citet{bender-friedman-2018-data} introduced data statements, a documentary framework for NLP datasets that seeks to provide a universal minimum bar for dataset description. Similar work has been done on systematizing documentation in other areas in data science and machine learning, including work focusing on 
online news~\citep{kevin-etal-2018-information}, data ethics~\citep{sun2019mithralabel}, and data exploration~\citep{holland2018dataset}, as well as generalist work such as~\citet{gebru2018datasheets}. 
Data quality is also implicitly documented by successes of filtering methods. There is a large literature on filtering data for various NLP tasks, e.g. \citet{axelrod-etal-2011-domain,moore-lewis-2010-intelligent,rarrick2011mt,wang-etal-2018-denoising,kamholz-etal-2014-panlex,junczys-dowmunt-2018-dual,caswell-etal-2020-language}. 

Closest to our work is the analysis of a highly multilingual (non-publicly available) web-crawl and LangID related quality issues by \citet{caswell-etal-2020-language}.
%, performing a highly multilingual web-crawl and then systematically analyzing the LangID related quality issues. 
%However, though 
They perform a brief analysis of the quality of OSCAR
%, but omit analyses of any other public datasets, 
with the focus only on the presence of in-language content.
\citet{dodge-c4} automatically documented and analyzed the contents and sources of C4~\citep{raffel2020exploring}, the English counterpart of mC4, which surfaced the presence of machine-translated contents and NLP benchmark data.

\section{Multilingual Corpora}\label{sec:crawls}
Table~\ref{tab:corpora} provides an overview of the corpora of interest in this work. We selected the corpora for their multilinguality and the inclusion of understudied languages in NLP. With the exception of WikiMatrix and ParaCrawl, all corpora are derived from CommonCrawl (CC).\footnote{\url{http://commoncrawl.org/}} %s, and distinguish themselves by the choice of filtering methods, LangID and automatic alignment technology.
%LangID is crucial to corpus creation, since any issues within might propagate, e.g. bias recognizing document types similar to what it was trained on, or errors in language identifiers.\footnote{\url{https://github.com/facebookresearch/fastText/issues/482}}

\paragraph{CCAligned~\citep{el-kishky-etal-2020-ccaligned}}is a %119-language\footnote{119 of originally 137 available for download (02/2021)}
%Although 137 language pairs are reported in ~\citet{el-kishky-etal-2020-ccaligned}, only 119 sentence-level corpora were available to download on \url{statmt.org} as of February 2021.}
parallel dataset built off 68 CC snapshots. Documents are aligned if they are in the same language according to FastText LangID~\citep{joulin-etal-2016-fasttext,joulin-etal-2017-bag}, and have the same URL but for a differing language code. These alignments are refined with cross-lingual LASER embeddings \citep{artetxe-schwenk-2019-massively}. For sentence-level data, they split on newlines and align with LASER, but perform no further filtering.
Human annotators evaluated the quality of document alignments for six languages (\texttt{de}, \texttt{zh}, \texttt{ar}, \texttt{ro}, \texttt{et}, \texttt{my}) selected for their different scripts and amount of retrieved documents, reporting precision of over 90\%.
% Latin, Chinese, Arabic, Burmese script
The quality of the extracted parallel sentences was evaluated in a machine translation (MT) task on six European (\texttt{da}, \texttt{cr}, \texttt{sl}, \texttt{sk}, \texttt{lt}, \texttt{et}) languages of the TED corpus~\citep{qi-etal-2018-pre}, where it compared favorably to systems built on crawled sentences from WikiMatrix and ParaCrawl v6. %, yielding BLEU scores in a range between 15 and 38.  

\paragraph{Multilingual C4 (mC4)~\citep{xue-etal-2021-mt5}} is a document-level dataset used for training the mT5 language model. It consists of monolingual text in 101 languages and is generated from 71 CC snapshots. It filters out pages that contain less than three lines of at least 200 characters and pages that contain bad words.\footnote{\url{https://github.com/LDNOOBW/}} Since this is a document-level dataset, we split it by sentence and deduplicate it before rating. For language identification, it uses CLD3~\citep{botha-etal-2017-natural},\footnote{\url{https://github.com/google/cld3/}} a small feed-forward neural network that was trained to detect 107 languages.
The mT5 model pre-trained on mC4
is evaluated on 6 tasks of the XTREME benchmark~\citep{hu2020xtreme} covering a variety of languages and outperforms other multilingual pre-trained language models such as mBERT~\citep{devlin-etal-2019-bert} and XLM\nobreakdash-R~\citep{conneau-etal-2020-unsupervised}.%\footnote{mBERT is trained on Wikipedia, {XML\nobreakdash-R} on CommonCrawl.} 

\paragraph{OSCAR~\citep{ortiz-suarez-etal-2019-asynchronous, ortiz-suarez-etal-2020-monolingual}}is a set of monolingual corpora extracted from CC snapshots, specifically from the plain text \emph{WET} format distributed by CC which removes all the HTML tags and converts the text to UTF-8. It is deduplicated and follows the approach by~\citep{grave-etal-2018-learning} of using FastText LangID~\citep{joulin-etal-2016-fasttext, joulin-etal-2017-bag} on a line-level.\footnote{\url{https://fasttext.cc/docs/en/language-identification.html} } No other filtering was applied.
For five languages (\texttt{bg}, \texttt{ca}, \texttt{da}, \texttt{fi}, \texttt{id}) OSCAR was used by its original authors to train language models which were then evaluated on parsing and POS tagging \citep{ortiz-suarez-etal-2020-monolingual}. OSCAR has also been used in independent studies to train monolingual or multilingual language models (\texttt{ar}, \texttt{as}, \texttt{bn}, \texttt{de}, \texttt{el}, \texttt{fr}, \texttt{gu}, \texttt{he}, \texttt{hi}, \texttt{kn}, \texttt{ml}, \texttt{mr}, \texttt{nl}, \texttt{or}, \texttt{pa}, \texttt{ro}, \texttt{ta}, \texttt{te}) and subsequently evaluate them on various downstream tasks \citep{antoun-etal-2021-araelectra, kakwani-etal-2020-indicnlpsuite, wilie-etal-2020-indonlu, chan-etal-2020-germans, koutsikakis-etal-2020-greek, martin-etal-2020-camembert, chriqui-etal-2021-hebert, seker-etal-2021-alephbert, delobelle-etal-2020-robbert, dumitrescu-etal-2020-birth, masala-etal-2020-robert}.

\paragraph{ParaCrawl v7.1} is a parallel dataset with 41 language pairs primarily aligned with English (39 out of 41) and mined using the parallel-data-crawling tool Bitextor \citep{espla-etal-2019-paracrawl,banon-etal-2020-paracrawl} which includes downloading documents, preprocessing and normalization, aligning documents and segments, and filtering noisy data via Bicleaner.\footnote{\url{https://github.com/bitextor/bicleaner}}
ParaCrawl focuses on European languages, but also includes 9 lower-resource, non-European language pairs in v7.1.
% TODO: https://www.aclweb.org/anthology/2020.eamt-1.31.pdf \citep{ramirez-sanchez-etal-2020-bifixer}
Sentence alignment and sentence pair filtering choices were optimized for five languages (\texttt{mt}, \texttt{et}, \texttt{hu}, \texttt{cs}, \texttt{de}) by training and evaluating MT models on the resulting parallel sentences. An earlier version (v5) was shown to improve translation quality on WMT benchmarks for~\texttt{cs}, \texttt{de}, \texttt{fi}, \texttt{lv}, \texttt{ro}.

\paragraph{WikiMatrix~\citep{schwenk-etal-2021-wikimatrix}} is a public dataset containing 135M parallel sentences in 1620 language pairs (85 languages) mined from Wikipedia. Out of the 135M parallel sentences, 34M are aligned with English. %, which we focus on.
%which relies on first learning multilingual sentence embeddings, and then applying the cosine distance metric to determine whether two sentences are close enough to be considered translations of each other.
The text is extracted from Wikipedia pages, split into sentences, and duplicate sentences are removed. FastText LangID is used before identifying bitext with LASER's distance-based mining approach. 
The margin threshold is optimized by training and evaluating downstream MT models on four WMT benchmarks (\texttt{de-en}, \texttt{de-fr}, \texttt{cs-de}, \texttt{cs-fr}). The final dataset is used to train translation models that are then evaluated by automatically measuring the quality of their translations against human translations of TED talks in 45 languages, with highest quality for translations between English and e.g. \texttt{pt}, \texttt{es}, \texttt{da}, and lowest for \texttt{sr}, \texttt{ja}, \texttt{mr}, \texttt{zh\_TW}.
In the audit we focus on language pairs with English on one side.
% https://github.com/facebookresearch/LASER/blob/master/tasks/WikiMatrix/WikiMatrix-bleu.pdf

\section{Auditing Data Quality}\label{sec:audit}
None of the above datasets has been evaluated for quality on the sentence level (exception: several languages in ParaCrawl v3), and downstream evaluations are centered around a small fraction of higher-resource languages. This is insufficient for drawing conclusions about the quality of individual or aligned sentences, and about the entirety of languages. In addition, there might be a publication bias preventing negative results with any of the above corpora with lower quality being published.

To close this gap, we conduct a human data quality audit focused on the lowest-resource and most under-evaluated languages, but also covering mid- and high-resource languages for comparison.

\begin{table*}[th]
    \small
    \centering
    \begin{tabular}{ll}
    \toprule
    	\multicolumn{2}{c}{\textbf{Correct Codes}} \\
    \midrule
      \textbf{\texttt{C}}:  \textit{Correct translation, any} & Combined label for \texttt{CC}, \texttt{CB}, \texttt{CS} \\
      \midrule
    	\multicolumn{2}{l}{\textbf{\texttt{CC}:} \textit{Correct translation, natural sentence}} \\
    %	\midrule
     \texttt{en} The Constitution of South Africa &  \texttt{nso} Molaotheo wa Rephabliki ya Afrika Borwa \\
     \texttt{en} Transforming your swimming pool into a pond & \texttt{de} Umbau Ihres Swimmingpools zum Teich \\
     \midrule
    	\multicolumn{2}{l}{\textbf{\texttt{CB}:} \textit{Correct translation, Boilerplate or low quality}} \\
    %	\midrule
     \texttt{en} Reference number: 13634 & \texttt{ln} Motango ya référence: 13634 \\
     \texttt{en} Latest Smell Stop Articles &	\texttt{fil} Pinakabagong mga Artikulo Smell Stop \\
    %  \texttt{en} Weight (Male): 8,6 - 13,5 kg & \texttt{ts} Ntiko (Xinuna): 8, 6 - 13, 5 kg \\
     \midrule
    	\multicolumn{2}{l}{\textbf{\texttt{CS}:} \textit{Correct translation, Short}} \\
   % 	\midrule
     \texttt{en} movies, dad & \texttt{it} cinema, pap\`{a} \\
     \texttt{en} Halloween - without me & \texttt{ay} Hallowen – janiw nayampejj \\
     \midrule
     \midrule
    	\multicolumn{2}{c}{\textbf{Error Codes}} \\
    	\midrule
\multicolumn{2}{l}{\textbf{X:} \textit{Incorrect translation, but both correct languages}} \\
    %	\midrule
     \texttt{en} A map of the arrondissements of Paris & \texttt{kg} Paris kele mbanza ya kimfumu ya Fwalansa.  \\
     \texttt{en} Ask a question & \texttt{tr} Soru sor Kullanıma g{\"o}re se\c{c}im \\
     \midrule
\multicolumn{2}{l}{\textbf{\texttt{WL}:} \textit{Source OR target wrong language, but both still linguistic content}} \\
   % 	\midrule
     \texttt{en} The ISO3 language code is zho & \texttt{zza} T{\' a}im eadra bracach mar bhionns na frogannaidhe. \\
    %  \texttt{en} sicilianu: Johannesburg & \texttt{ve} Tshivenda: Johannesburg \\
     \texttt{en} Der Werwolf — sprach der gute Mann, & 
 \texttt{de} des Weswolfs, Genitiv sodann, \\
     \midrule
\multicolumn{2}{l}{\textbf{NL:} \textit{Not a language: at least one of source and target are not linguistic content}} \\
    %	\midrule
     \texttt{en} EntryScan 4 \_ & \texttt{tn} TSA PM704 \_ \\
     \texttt{en} organic peanut butter & \texttt{ckb} \ucr \ucr \ucr \ucr \ucr \ucr \ucr \\
    \bottomrule
    \end{tabular}
    \caption{Annotation codes for parallel data with sentence pair examples. The language code before each sentence indicates the language it is supposed to be in.}
    \label{tab:examples}
\end{table*}

\subsection{Auditing Process} 

\paragraph{Participants} We recruited 51 volunteers from the NLP community, covering about 70 languages with proficient language skills.\footnote{This surprisingly high number comes in part because there are many closely related languages, e.g. one person may be proficient enough to rate many different Slavic or Turkic languages even if only one is their native language.} Each sentence is annotated by one rater.
To verify our hypothesis that those annotations can largely done by non-native speakers, we repeat a set of language expert annotations by a non-expert, and measure the accuracy of the non-expert. 

\paragraph{Sample selection} For each language in each dataset, we took a random sample of 100 lines, which may be anywhere from single words to short paragraphs depending on segmentation.
We manually annotated them according to the error taxonomy described below. For WikiMatrix and CCAligned, we selected those languages that are paired with English, and for ParaCrawl, we also included those paired with Spanish (``total'' counts in Table~\ref{tab:results}).
We did not annotate all languages, but focused on the ones with the least number of sentences in each dataset (at least the smallest 10) and languages for which we found proficient speakers. 
Since we annotate the same maximum number of sentences\footnote{Some languages had fewer than 100 sentences.} across all chosen languages regardless of their total number of sentences, the annotated samples are not an unbiased sample from the whole dataset. 

\paragraph{Non-expert labeling strategies}
Although many of the volunteers were familiar with the languages in question or spoke related languages, in cases where no speaker of a relevant language could be found, volunteers used dictionaries and internet search to form educated guesses. We discuss this deeper in Appendix~\ref{app:strategies} to highlight how much of this low-resource focused evaluation can actually be done by non-proficient speakers with relatively low effort.
In general, we aim to find an upper bound on quality, so we encouraged annotators to be forgiving of translation mistakes when the overall meaning of the sentence or large parts thereof are conveyed, or when most of the sentence is in the correct language. 

\paragraph{Effort} The individual effort was dependent on the quality and complexity of the data, and on the annotator's knowledge of the language(s), e.g., it took from less than two minutes for an English native speaker to pass through 100 well-formed English sentences (or similarly to annotate languages with 0\% in-language sentences), to two hours of ``detective work'' for well-formed content in languages for an annotator without familiarity.

\paragraph{Taxonomy}
In order to quantify errors, we developed a simple error taxonomy. Sentences and sentence pairs were annotated according to a simple rubric with error classes of Incorrect Translation (\texttt{X}, excluded for monolingual data), Wrong Language (\texttt{WL}), and Non-Linguistic Content (\texttt{NL}). Of correct sentences (\texttt{C}), we further mark single words or phrases (\texttt{CS}) and boilerplate contents (\texttt{CB}).
In addition, we asked annotators to flag offensive or pornographic content.
Table \ref{tab:examples} provides examples for parallel data, and Appendix~\ref{app:taxonomy} contains detailed annotation instructions.

\begin{table*}[th]
    \centering \small
   % \resizebox{\textwidth}{!}{%
    \begin{tabular}{llccccc}
    \toprule
   &&  \multicolumn{3}{c}{\textbf{Parallel}} & \multicolumn{2}{c}{\textbf{Monolingual}} \\
  \cmidrule(lr){3-5} \cmidrule(lr){6-7}
   & & \textbf{CCAligned} & \textbf{ParaCrawl v7.1} & \textbf{WikiMatrix} & \textbf{OSCAR} & \textbf{mC4}\\
    \midrule
    \multicolumn{2}{l}{\#langs audited / total} & 65 / 119 &  21 / 38 & 20 / 78 & 51 / 166  &  48 / 108 \\
    \multicolumn{2}{l}{\%langs audited} & 54.62\% &  55.26\% & 25.64\% & 30.72\% &  44.44\%\\
     \multicolumn{2}{l}{\#sents audited / total} & 8037 / 907M & 2214 / 521M   & 1997 / 95M & 3517 / 8.4B & 5314 / 8.5B \\
    \multicolumn{2}{l}{\%sents audited} & 0.00089\% & 0.00043\% & 0.00211\% & 0.00004\% & 0.00006\%\\
    \midrule
 \multirow{6}{*}{\rotatebox[origin=c]{90}{\textbf{macro}}}  &
 \texttt{C}       &     29.25\% &     76.14\% &      23.74\% & 87.21\% & 72.40\% \\
& \texttt{X}      &     29.46\% &     19.17\% &      68.18\% &  - &  - \\
& \texttt{WL}         &      9.44\% &      3.43\% &       6.08\% &  6.26\% & 15.98\% \\
& \texttt{NL}        &     31.42\% &      1.13\% &       1.60\% &  6.54\% & 11.40\% \\
& offensive &      0.01\% &      0.00\% &       0.00\% &  0.14\% &  0.06\% \\
& porn      &       5.30\% &      0.63\% &       0.00\% &  0.48\% &  0.36\% \\
    \midrule
\multirow{6}{*}{\rotatebox[origin=c]{90}{\textbf{micro}}} &  
\texttt{C}             &     53.52\% &     83.00\% &      50.58\% & 98.72\% & 92.66\% \\
& \texttt{X}         &     32.25\% &     15.27\% &      47.10\% &  - &  - \\
& \texttt{WL}        &      3.60\% &      1.04\% &       1.35\% &  0.52\% &  2.33\% \\
& \texttt{NL}       &      10.53\% &      0.69\% &       0.94\% &  0.75\% &  5.01\% \\
& offensive &      0.00\% &      0.00\% &       0.00\% &  0.18\% &  0.03\% \\
& porn      &      2.86\% &      0.33\% &       0.00\% &  1.63\% &  0.08\% \\
\midrule
& \#langs =0\% \texttt{C} & 7 & 0 & 1 & 7 & 0\\
%& \#langs $<$5\% \texttt{C} & 14 & 0 & 2 & 7  & 0 \\
%& \#langs $<$20\% \texttt{C} & 27 & 0 & 10 & 7 & 4 \\
& \#langs $<$50\% \texttt{C} & 44 & 4 & 19 & 11  & 9 \\
& \#langs $>$50\% \texttt{NL} & 13 & 0 & 0 & 7 & 1 \\
& \#langs $>$50\% \texttt{WL} & 1 & 0 & 0 & 3 & 4 \\
    \bottomrule
    \end{tabular}%
  %  }
    \caption{Averages of sentence-level annotations across datasets and selected languages. Macro-avg: Each language is weighted equally in the aggregation, regardless of its size. Micro-avg: Each label is weighted by the fraction of sentences for that language in the overall annotated corpus, i.e., the annotations for higher-represented languages are upweighted, and annotations for lower-represented languages are downweighted. The bottom rows contain the number of languages that have 0\% labeled \texttt{C} etc. Note that these are not true expectations since the languages audited were not randomly sampled. }
    \label{tab:results}
\end{table*}

\subsection{Human Audit Results}\label{sec:audit-res}

\paragraph{Interpretation of Results}
For each language, we compute the percentage of each label within the 100 audited sentences.
Then, we either aggregate the labels across languages with equal weights (macro-average), or weight them according to their presence in the overall dataset (micro-average). Results are shown in Table~\ref{tab:results}. The statistics for the correct codes (\texttt{CC}, \texttt{CB}, \texttt{CS}) are combined as \texttt{C}.
The number of languages, the numbers of sentences per language and the choice of languages differ across datasets, both in the original release and in the selection for our audit, so the comparison of numbers across datasets has to be taken with a grain of salt. Since the numbers are based on a small sample of sentences that were partially annotated by non-experts, the error statistics are only rough estimates.
Our audit captures a decent ratio of languages (25--55\%, 2nd row in Table~\ref{tab:results}), but only a tiny fraction of the overall number of sentences (0.00004--0.002\%).
When we speak of ``low-'' and ``high''-resource languages, we mean languages with smaller or larger representation in the datasets at hand. When reporting language-specific results we use the original language identifiers of the datasets.

\paragraph{Which datasets have quality issues?} 

The macro-averaged results show that the ratio of correct samples (\texttt{C}) ranges from 24\% to 87\%, with a large variance across the five audited datasets.
% Note that mC4 and OSCAR are monolingual datasets, so they do not require a correct alignment to another language to be labeled as correct.
Particularly severe problems were found in CCAligned and WikiMatrix, with 44 of the 65 languages that we audited for CCAligned containing under 50\% correct sentences, and 19 of the 20 in WikiMatrix. In total, 15 of the 205 language specific samples (7.3\%) contained not a single correct sentence. 
For the parallel datasets we are also interested in the quantity of misaligned/mistranslated sentences (\texttt{X}). For WikiMatrix, two-thirds of the audited samples were on average misaligned. We noticed that sentences were often similar in structure, but described different facts (see Table~\ref{tab:not_actually_parallel}). This might originate from the nature of the underlying Wikipedia articles, since they are often comparable rather than parallel~\citep{schwenk-etal-2021-wikimatrix}.

\begin{figure}[th!]
    \centering
    \includegraphics[width=\columnwidth]{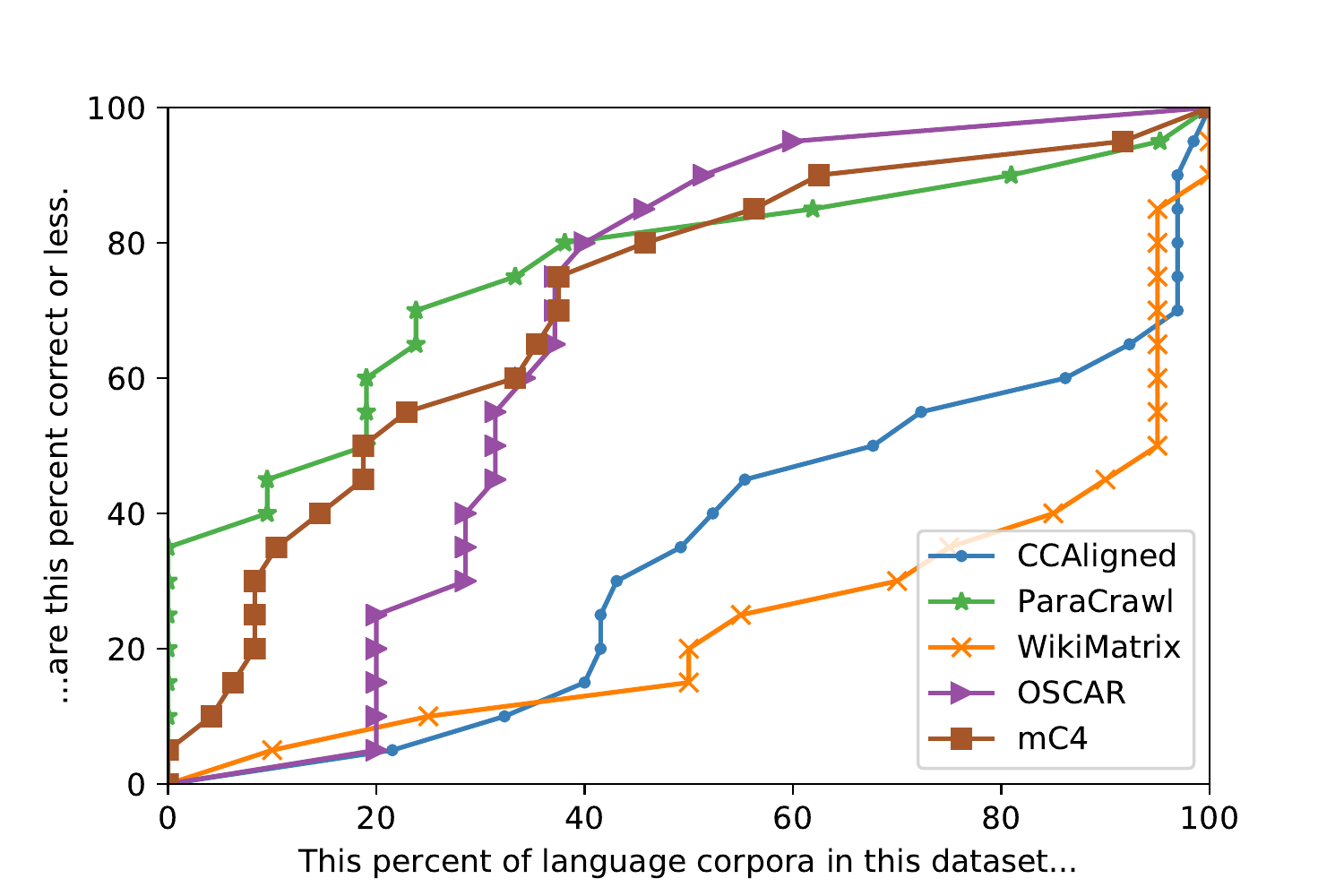}
    \caption{Fraction of languages in each dataset below a given quality threshold (percent correct).}% The larger the AUC, the better.}
    \label{fig:ratio_c}
\end{figure}

%While Table~\ref{tab:results} gives means and numbers of corpora passing certain thresholds, 
Figure~\ref{fig:ratio_c} illustrates per-corpus correctness more completely, showing for each dataset what percent of audited corpora are under each possible threshold of correctness.

\begin{figure*}[th]
\centering
\begin{subfigure}{.5\textwidth}
  \centering
  \includegraphics[width=\linewidth]{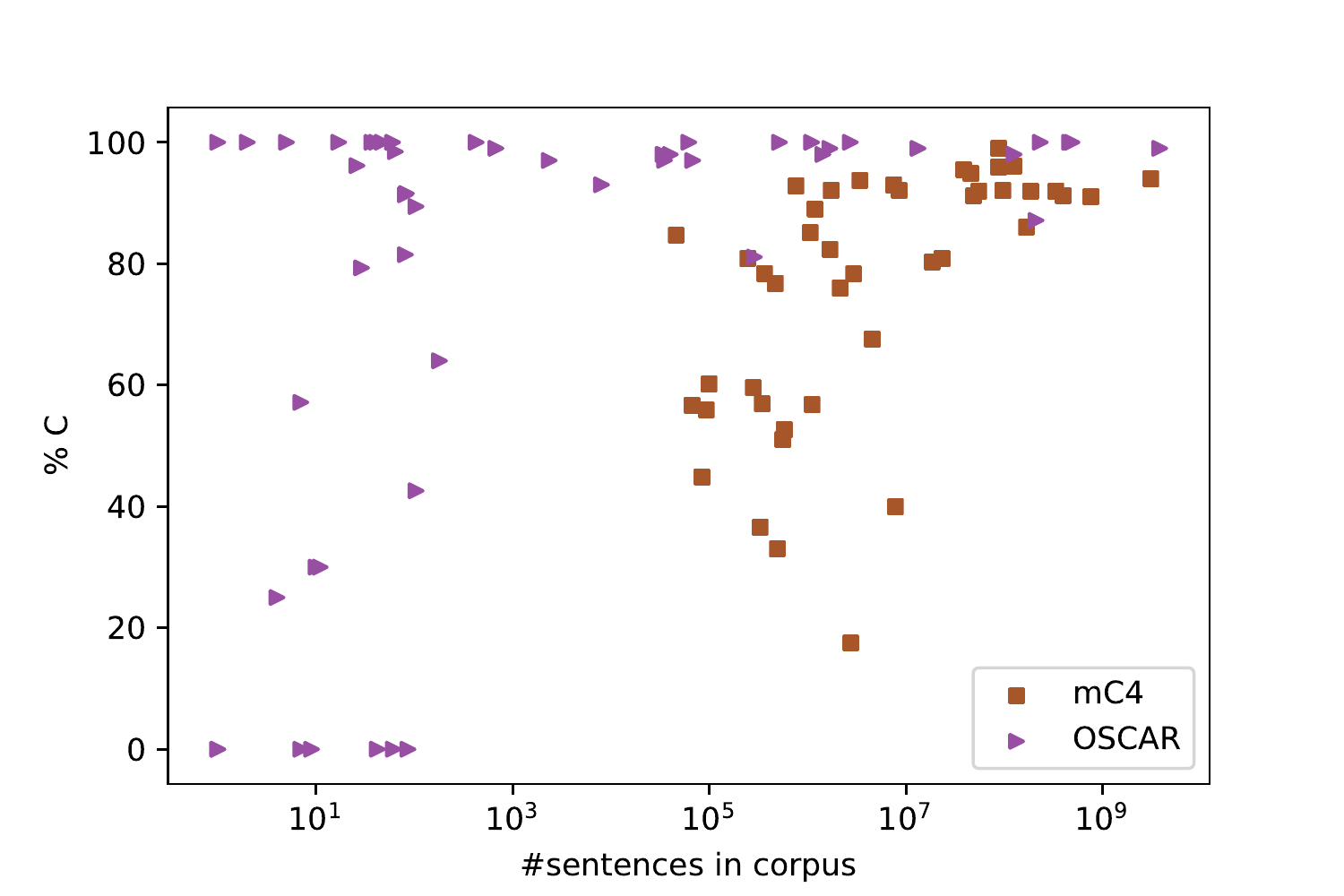}
  \caption{Monolingual corpora}
  \label{fig:C_mono}
\end{subfigure}%
\begin{subfigure}{.5\textwidth}
  \centering
  \includegraphics[width=\linewidth]{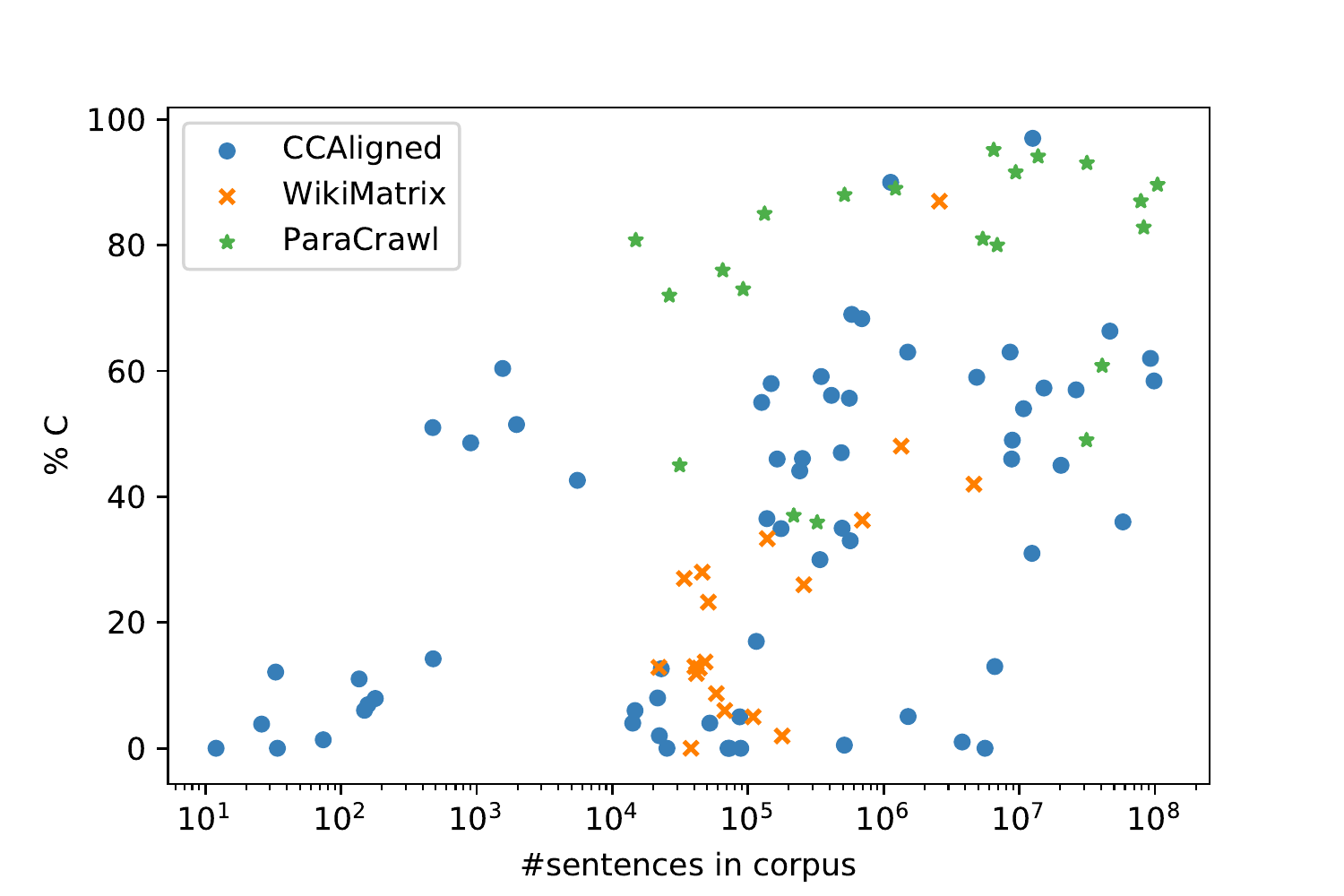}
  \caption{Parallel corpora}
  \label{fig:C_para}
\end{subfigure}
 \caption{Percentage of sentences labeled as correct vs. log N sentences for all audited languages.}
\label{fig:C}
\end{figure*}

\paragraph{Why haven't these problems been reported before?}
The findings above are averaged on a per-language basis (i.e. macro-average), and therefore give low and high-resource languages equal weight. If we instead estimate the quality on a per-sentence basis, i.e. down-weight lower-resource languages in the computation of the average, the numbers paint a more optimistic picture (``micro'' block in Table~\ref{tab:results}). This is especially relevant for the monolingual datasets because they contain audits for English, which makes up for 43\% of all sentences in OSCAR and 36\% in mC4. To illustrate the effect of this imbalance: A random sample from the entire mC4 dataset with over 63\% chance will be from one of the 8 largest languages (\texttt{en}, \texttt{ru}, \texttt{es}, \texttt{de}, \texttt{fr}, \texttt{it}, \texttt{pt}, \texttt{pl}, $>$100M sentences each), %\footnote{mC4 contains 22\% \texttt{und} sentences, i.e. sentences with undefined language.} 
 of which all have near perfect quality. Analogously, evaluation and tuning of web mining pipelines and resulting corpora in downstream applications focused largely on higher-resource languages (Section~\ref{sec:crawls}), so the low quality of underrepresented languages might go unnoticed if there is no dedicated evaluation, or no proficient speakers are involved in the curation~\citep{masakhane}.

\paragraph{How much content is nonlinguistic or in the wrong language?}
Nonlinguistic content is a more common problem than wrong-language content. Among the parallel datasets, CCAligned contains the highest percentage of nonlinguistic content, at 31.42\% on average across all rated corpora, and also the highest percent of wrong-language content, at 9.44\%. Among the monolingual datasets, mC4 contains the highest ratio both of sentences in incorrect languages (15.98\% average) and nonlinguistic content (11.40\% average), with 4 of the 48 audited languages having more than 50\% contents in other languages. The low amount of wrong language in ParaCrawl shows the benefits of selecting domains by the amount in-language text, but the dataset also covers the smallest amount of languages. The low ratio of wrong language samples in OSCAR may reflect the success of line-level LangID filtering.
These numbers provide evidence that more research in LangID could improve the overall quality, especially with respect to nonlinguistic content.

\paragraph{Which languages got confused?} The languages that were confused were frequently related higher-resource languages. However, there were also a significant number of ``out-of-model cousin" cases, where languages not supported by the LangID model ended up in a similar-seeming language. For instance in mC4, much of the Shona (\texttt{sn}, Bantu language spoken in Zimbabwe and Mozambique) corpus is actually Kinyarwanda (\texttt{rw}, Bantu language spoken in mostly in Rwanda and Uganda)---and, peculiarly, much of the Hawaiian (\texttt{haw}, Polynesian language spoken in Hawaii) is actually Twi (\texttt{tw}/\texttt{ak}, Central Tano language spoken mostly in Ghana).

\paragraph{Do low-resource languages have lower quality?}
Low-resource datasets tend to have lower human-judged quality.
The Spearman rank correlation between quality (\%\texttt{C}) and size is positive in all cases. The trend is strongest for mC4 ($r=0.66$), %,p=6.8e-8$), 
and gradually declines for CCAligned ($r=0.53$), %,p=0.0001$),
WikiMatrix ($r=0.49$), %,p=0.03$), 
ParaCrawl ($r=0.43$), %,p=0.02$) 
and OSCAR ($r=0.37$). %,p=0.10$).
Figure~\ref{fig:C} compares the number of sentences for each language against the proportion of correct sentences: %that we found during the audit: 
%The correlation between quality (\%\texttt{C}) and size is strongest for WikiMatrix (Pearson's $r=0.66$), while mC4, CCAligned, ParaCrawl have comparatively lower correlation (0.21, 0.25, 0.29), and OSCAR the lowest with $r=0.13$.
%In general, we observe that languages with low representation tend to contain fewer correct sentences, with an exception of a dozen of languages from OSCAR.
 Not all higher-resource languages ($>10^6$ sentences) have high quality, in particular for CCAligned (e.g. Javanese (\texttt{en\nobreakdash-jv\_ID}) with 5\%\texttt{C}, or Tagalog (\texttt{en\nobreakdash-tl\_XX}) with 13\%\texttt{C}). For mid-resource languages ($10^4$\nobreakdash--$10^6$ sentences) the picture is inconclusive, with some languages having high quality, and others having extremely low quality, even within the same datasets, e.g. Urdu in CCAligned \texttt{en-ur\_PK} has 100\%\texttt{C} vs. its romanized counterpart \texttt{en\nobreakdash-ur\_PK\_rom} 0.5\% \texttt{C}.
%\footnote{\texttt{\_rom} corpora have been removed in the latest CCAligned release.}
For individual error codes trends are less clear (not depicted).

\begin{table*}[!htbp]
    \centering
        \resizebox{\textwidth}{!}{%

    \begin{tabular}{lcccccccccccc}
        \toprule
        & \texttt{es\_XX} & \texttt{bm\_ML} & \texttt{yo\_NG} & \texttt{tr\_TR} & \texttt{ku\_TR} & \texttt{zh\_CN} & \texttt{af\_ZA} & \texttt{jv\_ID} & \texttt{zh\_TW} & \texttt{it\_IT} & \textbf{mean}\\
        \midrule
        \textbf{Acc-6} & 0.58 & 0.73 & 0.41 & 0.45 & 0.43 & 0.55 & 0.65 & 0.55 & 0.46& 0.55 & 0.66\\
        \textbf{Acc-4} & 0.77 & 0.73 & 0.60 & 0.55 & 0.56 & 0.72 & 0.72 & 0.57 & 0.58 & 0.66 & 0.72\\
        \textbf{Acc-2} & 0.91 & 0.96 & 0.72 & 0.64 & 0.71 & 0.79 & 0.77 & 0.92 & 0.81 & 0.69 & 0.79\\
         \bottomrule
    \end{tabular}%
     }
    \caption{Rater evaluation for a subset of audits from \textbf{CCAligned} (translated from English) measured by the accuracy (Acc-$n$) of annotations by non-proficient speaker against annotations by proficient speakers. 
    %$n$ indicates the granularity of the classes.  For $n=6$ all classes of the taxonomy were distinguished, for $n=4$ the \texttt{C} subclasses were combined, and for $n=2$ it is binary decision between \texttt{C} and the rest of the error classes.
    }
    \label{tab:agreement_ccaligned}
\end{table*}

\begin{table}[!htbp]
    \centering
      \resizebox{\columnwidth}{!}{%

    \begin{tabular}{lccccccccc}
        \toprule
        & \texttt{tyv} & \texttt{rm} &\texttt{bar} & \texttt{eml} & \texttt{zh} & \texttt{la} & \textbf{mean} \\
        \midrule
        \textbf{Acc-6} & 1.0 & 0.98 & 1.0 & 1.0 & 0.86 & 1.0 & 0.98 \\
        \textbf{Acc-4} &  1.0 & 1.0 & 1.0 & 1.0 & 0.87 & 1.0 & 0.98 \\
        \textbf{Acc-2} &  1.0 & 1.0 & 1.0 & 1.0 & 0.87 & 1.0 & 0.98 \\
         \bottomrule
    \end{tabular}%
    }
    \caption{Rater evaluation for a subset of audits from \textbf{OSCAR} measured by the accuracy (Acc-$n$) of annotations by non-proficient speaker against annotations by proficient speakers.}
    \label{tab:agreement_oscar}
\end{table}

\paragraph{Which languages have the lowest quality?} Across datasets we observe that the quality is particularly poor for languages that are included in romanized script (\texttt{\_rom}/\texttt{\_latn}), but are more commonly written in other scripts, e.g., Urdu (\texttt{ur}), Japanese (\texttt{ja}), Arabic (\texttt{ar}).
%\footnote{These romanized versions have been removed from CCAligned in a later release.} 
These are not transliterations of other scripts, but mostly contain non-linguistic material or wrong languages (e.g. the romanized Japanese corpus in mC4 (\texttt{ja\_latn}) contains Spanish, French, English, Portuguese, amongst others). %, Chinese (\texttt{zh}), Telugu (\texttt{te}) and Bulgarian (\texttt{bg}).  
In terms of geography, the poorest quality is found for African languages (Bambara (\texttt{bm}), Fula (\texttt{ff}), Kikongo (\texttt{kg}), Luganda (\texttt{lg}), Lingala (\texttt{ln}), Norther Sotho (\texttt{nso}), Oromo (\texttt{om}), Shona (\texttt{sn}), Somali (\texttt{so}), Tswana (\texttt{tn}), Wolof (\texttt{wo})), minority languages in Europe and the Middle East that are closely related to higher-resource languages (Azerbaijani (\texttt{az-IR}), North Frisian (\texttt{frr}), Neapolitan (\texttt{nap}), Silesian (\texttt{szl}), Zaza (\texttt{zza})), lesser spoken Chinese languages sharing a script with Mandarin (Yue (\texttt{yue}), Wu (\texttt{wuu})), four major Austronesian (Central Bikol (\texttt{bcl}), Chavacano (\texttt{cbk}), Javanese (\texttt{jv}), Sundanese (\texttt{su})), and some South-Asian languages, in particular Sinhala (\texttt{si}).
Appendix~\ref{app:stats} contains the detailed per-language statistics for all corpora. 
% Omitted from above: mt

\paragraph{What is the incidence of offensive and pornographic content?}
Overall, the sampled sentences did not contain a large amount of offensive contents. However, there were notable amounts of pornographic content ($>10\%$) found in CCAligned for 11 languages. % not fully annotated: tl_XX, lt_LV ?

\paragraph{Annotation quality}
For a subset of audited languages from CCAligned and OSCAR we measure the accuracy (Acc) of the labels assigned by non-proficient speakers against the labels assigned by proficient speakers for all audited sentences. This can be understood as a directed measure of annotator agreement for the special case where one rater is an expert and the other is not. Results for varying label granularity are reported in Tables~\ref{tab:agreement_ccaligned} and \ref{tab:agreement_oscar}. For $n=6$ all classes of the taxonomy were distinguished, for $n=4$ the \texttt{C} subclasses were combined, and for $n=2$ it is binary decision between \texttt{C} and the rest of the error classes. With the full 6-class taxonomy (Acc-6) we find a mean accuracy of 0.66 
%($\sigma^2 =0.02$) 
for CCAligned audits, and 0.98 
%($\sigma^2 =0.002$) 
for OSCAR audits. % (see appendix~\ref{app:agreement} for language-specific results).
With a binary taxonomy (Acc-2) distinguishing \texttt{C} from the rest, the accuracy further increases to 0.79 
%($\sigma^2=0.01$) 
for CCAligned. This provides strong evidence that good quality annotations are not limited to those proficient in a language. 

However, the significant drop of accuracy for finer-grained labels hints at that our taxonomy can be further improved, especially for parallel sentences. 
The error taxonomy lacks at least one category of error, namely ``correct/in-language but unnatural".  Similarly, the definition of ``correct-short" and ``correct-boilerplate" were not understood equally by all annotators and the concept of ``correct-short" has potential issues for agglutinative languages like Turkish. Finally, it was unclear what to do with related dialects, e.g. when a sentence is ``almost correct but wrong dialect" or when it is unclear which dialect a sentence belongs to. We recommend including these categories for future audits.

\subsection{Automatic Filtering}
Given the frequency of \texttt{WL} and \texttt{NL} annotations, it might be tempting to use open-source LangID models to post-filter data on a per-sentence(-pair) level, as OSCAR does. Unfortunately, this turns out to have its own issues.

\paragraph{Sentence-level n-gram LangID filtering}
We classify all sentence pairs of CCAligned with CLD3, an n-gram based LangID model. By comparing its predictions to the audit labels, we evaluate its quality on the subset of annotated samples: the classifier should detect both correct languages when the pair is annotated as \texttt{C} and \texttt{X}, and should detect incorrect languages in the pair when \texttt{WL} and \texttt{NL}. On this task, the CLD3 classifier
%\footnote{\texttt{filter=0.976 Prec, 0.962 Rec, 0.969 F1.}} 
achieves an average precision of only 40.6\%. %,
%n average accuracy of 56.4\% against our annotators across all audited sentences, 
%underlining the issues with LangID on web domain data~\citep{caswell-etal-2020-language}. %Its recall for detecting those pairs with wrong language(s) is 77.8\%, and its precision 35.9\%. 

\paragraph{Sentence-level Transformer LangID filtering}
N-gram LangID models like CLD3 have known problems. However, \citet{caswell-etal-2020-language} demonstrate that semi-supervised Transformer-based LangID models strongly out-perform them. We train a comparable Transformer-based LangID model and apply it to our annotated CCAligned data. We find that filtering noisy corpora ($<$ 50\% correct) on LangID for both source and target leads to gains in median precision, rising from 13.8\% pre-filter to 43.9\% post-filter. However, this comes at a steep cost of 77.5\% loss in recall. 
The biggest winners were Lingala, whose precision climbs from 8\% to 80\%, and Oromo, which soars from 2\% to 33\% in-language. Both of these, however, come at the cost of losing 50\% of the correct in-language sentences, being reduced from ~22k sentences to 3k and 1k sentences respectively, which would likely be too small for building downstream models. The moral is that, at least at the current stage, there is no one-size-fits-all approach for sentence-level LangID filtering.

\section{Dataset Mis-labeling}
\label{sec:codes}
Standardized and unambiguous representations of language codes are important for practical data use and exchange. The standard used by most academic and industry applications is BCP-47~\citep{phillips2006tags}, which builds off the two-letter ISO639-2 codes and three-letter ISO639\nobreakdash-3 codes, but also allows to add subtags for scripts (e.g. Hindi in Latin script: \texttt{hi-Latn}) or regional varieties (e.g. French spoken in Canada: \texttt{fr-CA}). It would enhance transparency and interoperability if adopted consistently, especially with growing language diversity in NLP. % since it allows to add subtags for scripts or regional varieties. 
%, which builds off the two-letter ISO639-2 codes and three-letter ISO639\nobreakdash-3 codes. Codes may additionally specify ISO15924 script subtags to indicate that a nonstandard script is used (e.g. \texttt{hi-Latn} for Hindi written in Latin script), ISO3166-1 country codes to indicate regional varieties (e.g. \texttt{fr-CA} for Canadian French), or extensions for private use (e.g. \texttt{ca-x-val} for Valencian Catalan). Some BCP-47 codes represent groups of languages---for instance, \texttt{kg} represents the Kongo language, and \texttt{kng}, \texttt{ldi}, \texttt{kwy}, and \texttt{yom} represent particular varieties of Kongo.

We find a variety of errors and inconsistencies in language code usage, ranging from serious mislabelings to small transgressions against standard conventions. For this analysis, we also include the JW300~\citep{agic-vulic-2019-jw300} dataset, a multilingual dataset crawled from \url{jw.org}. %, which was otherwise not audited in this paper. 
In summary, we find 8 nonstandard codes in CCAligned, 3 in OSCAR, 1 in mC4, 1 in WikiMatrix, and 70 in JW300, for 83 in total. This does not include the 59 codes affected by superset issues. %0 in ParaCrawl, 
Full details are given in Appendix~\ref{app:jw300}.

\paragraph{Inconsistent Language Codes} One common issue is simply using nonstandard or invented codes. For example, CCAligned uses only two-letter codes, so when the BCP-47 code for a language is three letters it is either shortened (e.g. \texttt{zza} $\rightarrow$ \texttt{zz})
%, \texttt{szl}  $\rightarrow$ \texttt{sz}, \texttt{nso}  $\rightarrow$ \texttt{ns}, \texttt{ckb}  $\rightarrow$ \texttt{cb}, \texttt{ber}  $\rightarrow$ \texttt{tz} \footnote{Tamazight (BCP-47 ber) goes by various codes, so this may have been a shortening of e.g. \texttt{tzm}}) 
or invented (\texttt{shn}  $\rightarrow$ \texttt{qa}).
%, \texttt{kac}  $\rightarrow$ \texttt{qd}, \texttt{ceb}  $\rightarrow$ \texttt{cx}), which can lead to 
%this can lead to confusion and limits the compatibility with other tools and resources.
Similarly, OSCAR contains data labeled as \texttt{als} (BCP-47 for Tosk Albanian) that is actually in \texttt{gsw} (Allemannic).\footnote{This is a result of the language code used by the \href{https://en.wikipedia.org/wiki/Alemannic\_Wikipedia}{Alemannic Wikipedia} and affects any corpus or tool that uses Wikipedia data without correcting for this, like FastText.} 
% \citep{joulin-etal-2017-bag, joulin-etal-2016-fasttext}.
%\footnote{\url{https://en.wikipedia.org/wiki/Alemannic_Wikipedia}
% And in JW300~\citep{agic-vulic-2019-jw300}, a multilingual dataset otherwise not audited in this paper, there are five codes ({\tt cat}, {\tt daf}, {\tt que}, {\tt nya}, {\tt run}) which use ISO693-3 instead of BCP-47 codes. Puzzlingly, in most of these cases there exist separate datasets for the equivalent ISO693-2 and 3 codes (e.g. both {\tt ny} and {\tt nya}).
22 additional language codes in JW300 have similar issues, 
%mostly from mis-parsed private-use extensions, 
including 12 codes that start with \texttt{jw\_} but are not %(as they may appear)
Javanese. 

\paragraph{False Sign Languages}
12\% (48/417) of JW300 
%has a much stranger problem than nonstandard codes. It has the peculiar issue that a full 12\% (48/417) of the languages it claims to cover 
carry language codes for sign languages. %While it is possible to transcribe sign languages using glosses, this is not what these corpora are. 
Instead of sign language transcripts they are texts in another high resource language, mostly English or Spanish---for example, the \texttt{en-zsl} (Zambian sign language) data is actually English-English parallel data (copies), details in Appendix~\ref{app:jw300}. This was likely caused by videos with sign language interpretation embedded on the crawled websites.\footnote{Kudos to Rebecca Knowles for this explanation.} %Details are in Appendix Table~\ref{tab:signlanguages}.

\paragraph{Mysterious supersets} 
When datasets contain language codes that are supersets of other language codes, it is difficult to determine which particular language the text contains. WikiMatrix has Serbian (\texttt{sr}), Croatian (\texttt{hr}), Bosnian (\texttt{bs}), and Serbo-Croatian (\texttt{sh})---their superset.\footnote{\url{https://iso639-3.sil.org/code/hbs}}
%. And while there may be some debate whether \texttt{bs},  \texttt{hr},  \texttt{cnr},  and \texttt{sr} are different languages, \texttt{sh} (\texttt{hbs}) is by definition a superset of all of them.\footnote{https://iso639-3.sil.org/code/hbs} 
The issue of codes that are supersets of others is common enough to include a small table dedicated to it (Appendix Table~\ref{tab:supersets}). 
In some cases this may not be an issue, as with Arabic, where \texttt{ar} conventionally refers to Modern Standard Arabic, even though the code technically encompasses all dialects.
%, or where \texttt{no} typically refers to Norwegian Bokm\r{a}l (\texttt{nb}), though it technically is the superset of \texttt{nb} and \texttt{nn}. 
In many cases, the nature of the data in the superset code remains a mystery.
% requiring detective work.

\paragraph{Deprecated codes} Finally, there are several deprecated codes that are used: \texttt{sh} in WikiMatrix, \texttt{iw} in mC4, \texttt{sh} and \texttt{eml} in Oscar, and \texttt{daf} in JW300.

\section{Risks of Low-Quality Data}\label{sec:risk}

\paragraph{Low quality in downstream applications}
Text corpora today are building blocks for many downstream NLP applications like question answering and text summarization---for instance, a common approach is to first train translation models on such data and then automatically translate training data for downstream models~\citep{conneau-etal-2018-xnli}. If the data used for the original systems is flawed, derived technology may fail for those languages far down the line without knowing the causes.
This risk of undesired downstream effects calls for future studies with a careful treatment of intertwined effects such as data size and domain, language-specific phenomena, evaluation data and metric biases.
%Furthermore, there are not many existing public models trained on these specific subsets of data that we can analyze. 
To give the reader a brief glimpse of the impact of data quality for the example of translation, we compare the \texttt{C}\% metric from our audit with the translation quality (sentencepiece-BLEU, spBLEU) of the multilingual translation model M2M124 for 124 languages~\citep{goyal-etal-2021-flores-101}. It was trained on WikiMatrix and CCAligned, and similar data collected with the same tools, which we expect to show similar biases. Translation quality is evaluated on the trusted, human-translated FloReS benchmark~\citep{goyal-etal-2021-flores-101}. 
%For language pairs that were both covered in the WikiMatrix and the CCAligned audit, we compute an average of their \% \texttt{C} scores weighted by their size. 
For the 21 languages present in both the audit and the FloReS benchmark, we found a positive correlation (Spearman) between the data quality scores and spBLEU of $\rho=0.44$ $(p=0.041)$. This is not as large as the correlation with data size ($\rho=0.66$, $p=0.00078$), but it nonetheless helps to explain translation quality---the correlation between the product of \texttt{C}\% and data size (in other words, the expected total number of good sentences in the dataset), is the highest yet, with a value of $\rho=0.73$ $(p=0.00013)$.\footnote{For the translation from English, BLEU scores are less comparable but the trend holds nonetheless, with values of ($\rho=0.32$, $p=0.14$), ($\rho=0.74$, $p=0.000078$), and ($\rho=0.80$, $p=0.0000087$) respectively.}
% The human inspection and auditing of e.g. trained vector representations to detect possible risks and misrepresentations of a subset of languages is arguably harder than manually inspecting a few samples as we did in this work.
% As our analysis has shown, low-resource languages are disproportionately affected by such problems in automatic data curation pipelines.

\paragraph{Representation washing}
Since there are datasets which contain many low-resource languages, the community may feel a sense of progress and growing equity, despite the actual quality of the resources for these languages. %However, models often still perform poorly on NLP tasks for these languages
%Because there appear to be datasets for low-resource languages, the community may collectively feel as though progress is being made in these areas. 
Similarly, if low-quality datasets are used as benchmarks they may exaggerate model performance, making low-resource NLP appear more solved than it is---or conversely, if models perform poorly when trained with such data, it may be wrongly assumed that the task of learning models for these languages is harder than it actually is or infeasible given current resources. These effects could result in productive effort being redirected away from these tasks and languages.
%The result can be that productive effort will be directed away from these fields.

\begin{table}[t!]
\resizebox{\columnwidth}{!}{%

\centering
\begin{tabular}{ll}
\toprule

\texttt{en} & The prime minister of the \textbf{UK} is \textbf{Boris Johnson}. \\
\texttt{nl} & De minister-president van \textbf{Nederland} is \textbf{Mark Rutte}. \\
& \small{\texttt{en}: The prime minister of the Netherlands is Mark Rutte.} \\
\midrule
%\midrule
% \texttt{en} &Sunglasses \\
% \texttt{ig}	&ah\d{i}a Nyocha \\
% \midrule
\texttt{en} &\textbf{24 March} 2018 \\
\texttt{pt}& 	\textbf{14 Novembro} 2018 \\
 & \small{\texttt{en}: 14 November 2018 }\\
%\midrule
\midrule
% \texttt{en} &The current local time in \textbf{Sarasota} is \textbf{89} minutes ahead of apparent solar time. \\
% \texttt{nn}	&Den lokale tiden i \textbf{Miami} er \textbf{86} minutt f\o{o}re sann soltid. \\
\texttt{en} &The current local time in \textbf{Sarasota} is \textbf{89} minutes. \\
\texttt{nn}	&Den lokale tiden i \textbf{Miami} er \textbf{86} minutt. \\
 & \small{\texttt{en}: The local time in Miami is 86 minutes.}\\
%\midrule
\midrule
\texttt{en} &In \textbf{1932} the highway was extended \textbf{north to LA}. \\
\texttt{bar} &\textbf{1938} is de Autobahn bei \textbf{Inglstod} fertig gstellt.	\\
 & \small{\texttt{en}: The highway near Inglstod was completed in 1938.}\\
% \midrule
% \textit{en:} He was engaged to the lawyer and actor João Lima Junior, with whom he dated from 2004 to 2006. \\
% \textit{nds:} Ze woont in de tussentied samen met zanger en liedtiesschriever Johannes Oerding met wie ze sinds 2009 ook samen op de bühne stiet.\\
\bottomrule
\end{tabular}%
}
\caption{Examples of ``parallel" data where the translation has a different meaning than the source, but the form looks the same. (We added translations of the non-English side.) Such data may encourage hallucinations of fake ``facts".}
\label{tab:not_actually_parallel}
\end{table}

\paragraph{Trust in incorrect ``facts''} % and trust} %, algorithmic trust and automation bias}
We found many instances of parallel-looking sentences that are structurally and semantically similar, but not factually correct translations (Table~\ref{tab:not_actually_parallel}). They can cause models to produce plausible ``translations" that are factually wrong, but users may still trust them (\textit{algorithmic trust}) without verifying the information. %This is relevant for \textit{algorithmic trust}, when users increasingly trust the outputs of computers and ``algorithms" without verifying the information. 
Similarly, \textit{automation bias} \citep{skitka1999does}, 
%from social psychology which refers to the bias of 
referring to humans favoring decisions made by automated systems over decisions made by humans, might amplify the issues of inaccurate translations caused by misalignments.
%One variant of this issue that occurs frequently in some datasets is pornographic content.
%, which in the majority of the cases we observed were parts of misaligned sentence pairs. 
%Another effect is that models trained on misaligned pornographic content may hallucinate such content, which may be disturbing to users.

\section{Future Work and Recommendations}\label{sec:recommendation}
Of the five multilingual corpora evaluated, we consistently found severe issues with quality, especially in the lower-resource languages. We rated samples of 205 languages, and found that 87 of them had under 50\% usable data, with a full 15 languages at 0\% in-language. We furthermore found consistent issues with mislabeled data and nonstandard language codes, particularly in the JW300 dataset, and identified 83 affected corpora, at least 48 of which were entirely spurious (Section~\ref{sec:codes}). While there might have been anecdotal evidence of insufficient quality for some of the datasets, the majority of these quality issues had not been reported, nor been investigated in depth. These issues might go unnoticed for languages that are not represented in the evaluation of the crawling methods, and cause harm in downstream applications~\citep{khayrallah-koehn-2018-impact}.

There are a variety of ways to improve both the ease and accuracy of human evaluation, as well a few classes of issues we ignored in this paper, like close dialects. 
Ideally we would like to build a standard suite of automatic metrics for datasets, but more research is necessary to determine what the appropriate metrics would be. One important area missing from our analyses however is the estimated portion of a dataset which has been generated by MT~\citep{rarrick2011mt}, LM systems, or bots/templates, as for example in the analysis of C4~\citep{dodge-c4}. %A prominent example is the Lsjbot\footnote{\url{https://en.wikipedia.org/wiki/Lsjbot}} which is responsible for creating 80-90\% of content for Swedish, Cebuano and Waray Wikipedia. 
The information captured in machine-generated content might still be useful for modeling, but might falsely overrepresent typical generation patterns and introduce linguistic errors or unnatural artifacts. 
% Malagasy wiktionary audit: https://meta.wikimedia.org/wiki/Requests_for_comment/Large-scale_errors_at_Malagasy_Wiktionary
%https://www.vice.com/en/article/4agamm/the-worlds-second-largest-wikipedia-is-written-almost-entirely-by-one-bot

% Finally, similar studies to this in future would do well ton work more on calibrating human raters, to ensure consistent use of error categories.

% An issue that arises with the progress in building technology for some of the languages is the retrieval of machine-generated output as in-language data. This is prominent for mid- to high-resource languages for which translation systems have reached sufficient quality for website translation, as we observed for example a significant amount of translations for Ukrainian. Non-native speakers might not have noticed them during annotations, so the problem might be even larger than we might estimate now. We leave a systematic investigation to future work. A more unexpected artifact is the retrieval of published BPE vocabularies for a range of low-resource languages, such as Sundanese.\footnote{\url{https://nlp.h-its.org/bpemb/su/su.wiki.bpe.vs100000.vocab}}

We therefore strongly recommend looking at samples of any dataset before using it or releasing it to the public. As we have shown, one does not need to be proficient in a language to see when there are serious quality issues, and a quick scan of 100 sentences can be sufficient to detect major problems. Moreover, going through and annotating a small sample of data can bring actionable insights about new ways to filter or use it.

If data quality issues are found, a wide variety of techniques can be explored, like filtering on length-ratio, LangID, TF-IDF wordlists \cite{caswell-etal-2020-language} or dictionaries~\citep{kamholz-etal-2014-panlex}; to neural approaches like LM scoring \cite{axelrod-etal-2011-domain,moore-lewis-2010-intelligent,wang-etal-2018-denoising}. Unfortunately, none of these provides a quick and easy fix, especially for low-resource languages---data cleaning is no trivial task!

Noisy datasets are by no means useless, at least if they contain some desirable content. Therefore an alternative to filtering can be documentation~\citep{bender2021dangers}. This can take the form of a per-language quality score and notes about known issues,
% ({ \it``language xx has high percentage non-linguistic content'' } etc.), 
a datasheet \citep{gebru2018datasheets} or nutrition label \citep{holland2018dataset}. However, we suggest researchers not release corpora with near-zero in-language content, as this may give the mistaken impression of usable resources.

Finally, we encourage the community to continue conducting evaluations and audits of public datasets---similar to system comparison papers.

\section*{Acknowledgements}
We would like to thank the TACL editors and reviewers, and AfricaNLP and Google reviewers who have helped us shape this paper. Furthermore, we are grateful for Ahmed El-Kishky's support and help with CCAligned and WikiMatrix size statistics.

\bibliography{anthology,custom}
\bibliographystyle{acl_natbib}

\appendix

\clearpage

\begin{table}[th!]
    \centering
\begin{tabular}{lll}
\toprule
 \textbf{Dataset } &    \textbf{Supercode} &  \textbf{Subcode(s)} \\
 \midrule
JW300 & \texttt{kg} & \texttt{kwy} \\                       
JW300 & \texttt{mg} & \texttt{tdx} \\
JW300 & \texttt{qu} & \texttt{que}, \texttt{qug}, \texttt{qus}, \\
& & \texttt{quw}, \texttt{quy}, \texttt{quz}, \\
& &  \texttt{qvi}, \texttt{qvz} \\
JW300 & \texttt{sw} & \texttt{swc} \\
 \midrule
OSCAR & \texttt{ar} & \texttt{arz} \\
OSCAR & \texttt{az} & \texttt{azb} \\
OSCAR & \texttt{sh} & \texttt{bs}, \texttt{hr}, \texttt{sr} \\
OSCAR & \texttt{ku} & \texttt{ckb} \\       
OSCAR & \texttt{ms} & \texttt{id}, \texttt{min} \\
OSCAR & \texttt{no} & \texttt{nn} \\
OSCAR & \texttt{sq} & \texttt{als}$^{*}$ \\
OSCAR & \texttt{zh} & \texttt{yue}, \texttt{wuu} \\
%  \midrule
% Tatoeba & ar & acm,afb,ajp,apc,arq,ary,arz,ayl \\
% Tatoeba & ber & kab \\
% Tatoeba & et & vro \\
% Tatoeba & ku & ckb,kmr,sdh \\
% Tatoeba & lv & ltg \\
% Tatoeba & sq & aln \\
 \midrule
WikiMatrix & \texttt{ar} & \texttt{arz} \\
WikiMatrix & \texttt{sh} & \texttt{bs}, \texttt{hr}, \texttt{sr} \\
WikiMatrix & \texttt{zh} & \texttt{wuu} \\    
\bottomrule
\end{tabular}
    \caption{Situations where two language codes are represented, but one is a superset of another by the ISO standard, leading to unclarity about the data in the supercode dataset. $^{*}$The \texttt{als} dataset is actually in \texttt{gsw}.}
    \label{tab:supersets}
\end{table}

\begin{table}[th!]
    \centering
    \small
\begin{tabular}{ll}
\toprule
\textbf{ Actual language}  &   \textbf{Code in JW300} \\
 \midrule
\texttt{cs} & \texttt{cse} \\
\texttt{de} & \texttt{gsg} \\
\texttt{el} & \texttt{gss} \\
\texttt{en} & \texttt{ase}, \texttt{asf}, \texttt{bfi}, \texttt{ins}, \texttt{psp}, \\
& \texttt{sfs}, \texttt{zib}, \texttt{zsl} \\
\texttt{es} & \texttt{aed}, \texttt{bvl}, \texttt{csf}, \texttt{csg}, \texttt{csn},\\
   &\texttt{csr}, \texttt{ecs}, \texttt{esn}, \texttt{gsm}, \texttt{hds}, \\
   & \texttt{lsp}, \texttt{mfs}, \texttt{ncs}, \texttt{prl}, \texttt{pys},\\
   & \texttt{ssp}, \texttt{vsl} \\
\texttt{fi} & \texttt{fse} \\
\texttt{fr} & \texttt{fcs},\texttt{fsl} \\
\texttt{hu} & \texttt{hsh} \\
\texttt{id} & \texttt{inl} \\
\texttt{it} & \texttt{ise} \\
\texttt{ja} & \texttt{jsl} \\
\texttt{ko} & \texttt{kvk} \\
\texttt{pl} & \texttt{pso} \\
\texttt{pt} & \texttt{bzs}, \texttt{mzy}, \texttt{psr}, \texttt{sgn\_AO} \\
\texttt{ro} & \texttt{rms} \\
\texttt{ru} & \texttt{rsl} \\
\texttt{sk} & \texttt{svk} \\
\texttt{sq} & \texttt{sql} \\
\texttt{st} & \texttt{jw\_ssa} \\
\texttt{zh} & \texttt{csl}, \texttt{tss} \\  
\bottomrule
\end{tabular}
    \caption{There are 48 languages in the JW300 corpus with language codes that correspond to sign languages, but in reality are unrelated high-resource languages (usually the most spoken language in the country of origin of the sign language). This table shows the actual language of the data corresponding to each sign language code.} %  For instance, the \texttt{ase-en} parallel data is actually \texttt{en-en} parallel data (copied source and target).
    \label{tab:signlanguages}
\end{table}

\section{Details on Language Code Issues}
\label{app:jw300}

Table \ref{tab:supersets} provides a complete lists of the corpora where one code is defined as a superset of the other by the ISO standard, and in Table \ref{tab:signlanguages} we provide a complete list of the language codes in JW300 which purport to be sign language but are actually unrelated high-resource languages.

\begin{table}[th!]
    \centering
     \resizebox{\columnwidth}{!}{%
\begin{tabular}{lll}

\toprule
 \textbf{Code in JW300} &  \textbf{BCP-47 code} & \textbf{Actual Language Name} \\
\multicolumn{3}{c}{} \\
\multicolumn{3}{c}{\textbf{Incorrect private-use extensions}} \\		
\midrule		
hy\_arevmda & 	hyw & 	Western Armenian \\
jw\_dgr & 	os\_x\_dgr & 	Digor Ossetian \\ 
jw\_dmr & 	naq\_x\_dmr & 	Damara Khoekhoe \\ 
jw\_ibi & 	yom\_x\_ibi & 	Ibinda Kongo \\ 
jw\_paa & 	pap\_x\_paa & 	Papiamento (Aruba) \\ 
jw\_qcs & 	qxl & 	Salasaca Highland Kichwa \\ 
jw\_rmg & 	rmn\_x\_rmg & 	Greek Romani (South) \\ 
jw\_rmv & 	rmy\_x\_rmv & 	Vlax Romani, Russia \\ 
jw\_spl & 	nso\_x\_spl & 	Sepulana \\ 
jw\_ssa & 	st\_ZA & 	Sesotho (South Africa) \\ 
jw\_tpo & 	pt\_PT & 	Portuguese (Portugal) \\ 
jw\_vlc & 	ca\_x\_vlc & 	Catalan (Valencia) \\ 
jw\_vz & 	skg\_x\_vz & 	Vezo Malagasy \\  
rmy\_AR & rmy\_x\_? & Kalderash \\ 

\multicolumn{3}{c}{} \\
\multicolumn{3}{c}{\textbf{Equivalent codes used in place of extensions}} \\		
\midrule	
kmr\_latn & 	kmr\_x\_rdu & 	Kurmanji (Caucasus) \\ 
nya & 	ny\_x\_? & 	Chinyanja (Zambia) \\ 
que & 	qu\_x\_? & 	Quechua (Ancash) \\ 

\multicolumn{3}{c}{} \\
\multicolumn{3}{c}{\textbf{Deprecated codes}} \\		
\midrule		
daf & 	dnj/lda & 	Dan \\ 
% sgn\_AO & 	pt & 	Portuguese \\ 

\multicolumn{3}{c}{} \\	
\multicolumn{3}{c}{\textbf{ISO-693-3 used in place of ISO-693-2}} \\		
\midrule		
cat & 	ca & 	Catalan \\ 
gug & 	gn & 	Guarani \\ 
run & 	rn & 	Kirundi \\ 
tso\_MZ & 	ts\_MZ & 	Changana (Mozambique) \\ 
\bottomrule
\end{tabular}%
}
    \caption{Language code issues in the JW300 datasets for 22 language varieties not covered by Tables \ref{tab:supersets} and \ref{tab:signlanguages}. 
    %Twelve languages have codes starting in \texttt{jw\_}, suggesting they are varieties of Javanese, but are instead mis-parsed private-use extensions. Three codes appear in addition to equivalent ISO codes, making it unclear which languages they are. One language uses a deprecated ISO code. Four languages use the ISO639-3 code instead of the ISO639-2 code, and therefore are not BCP-47. (Note: in this table, 
    Private use extensions are given as they appear in \url{jw.org}, and specified as `?' if they are absent from \url{jw.org}.}
    \label{tab:jw300nonbcp}
\end{table}

Special attention needs to be given to the JW300 dataset, which, in addition to the sign languages and superset code issues, has a variety of other peculiarities. These problems seem to originate in the codes used by \url{jw.org},\footnote{The \url{jw.org} website seems to use correct BCP-47 extensions now, however, and entering a code such as ``jw\_dmr" redirects to ``naq\_x\_dmr".} which were apparently not checked in the creation of the JW300 dataset. An overview is provided in Table \ref{tab:jw300nonbcp}, and the following paragraphs give specifics.

Twelve languages in JW300 have codes starting in \texttt{jw\_}, suggesting they are varieties of Javanese (ISO639-1 \texttt{jw}), but are instead attempts to represent language dialects for which there are no BCP-47 codes. These codes seem to have been updated in \url{jw.org} to appropriate BCP-47 private-use extensions in the form \texttt{<supercode>\_x\_<tag>}, which are provided in Table \ref{tab:jw300nonbcp}.
Twelve languages have codes starting in \texttt{jw\_}, suggesting they are varieties of Javanese, but are instead mis-parsed private-use extensions. Three codes appear in addition to equivalent ISO codes, making it unclear which languages they are. One language uses a deprecated ISO code. Four languages use the ISO639-3 code instead of the ISO639-2 code, and therefore are not BCP-47. 

In addition to the \texttt{jw\_} tags, there are two other mis-used private subtags: \texttt{hy\_arevmda}, which in addition to lacking the mandatory \texttt{\_x\_} appears to represent standard Western Armenian (\texttt{hyw}); and \texttt{rmy\_AR}, which, rather than being Romany from Argentina, is Kalderash Romany.

There are also a few anomalies where private use extensions should have been used but other methods were found to convey the distinctions. Three codes appear in addition to equivalent ISO codes, making it unclear which languages they are. Two of these are equivalencies between  ISO639-2 and  ISO639-3 (\texttt{nya} and \texttt{ny} are both Chichewa, \texttt{qu} and \texttt{que} are both Quechua), and one is a script equivalency (\texttt{kmr} and \texttt{kmr\_latn} are both in Latin script). In these three cases the two codes do represent different languages---so a private use extension would have been appropriate.

Finally, there is the more minor issue that three languages use the ISO639-3 code instead of the ISO639-2 code, and therefore are not BCP-47.

In addition to the JW300-specific errors, Table \ref{tab:misc_codes} summarizes miscellaneous errors in CCAligned and OSCAR that were detailed in Section \ref{sec:codes}.

\begin{table}[!th]
\small
\centering
\begin{tabular}{lll}
\toprule
\textbf{Dataset} & \textbf{Code in Corpus} & \textbf{Correct Code} \\
\midrule
CCAligned & \texttt{zz} & \texttt{zza} \\
CCAligned & \texttt{sz} & \texttt{szl} \\
CCAligned & \texttt{ns} & \texttt{nso} \\
CCAligned & \texttt{cb} & \texttt{ckb} \\
CCAligned & \texttt{tz} & \texttt{ber} \\
CCAligned & \texttt{qa} & \texttt{shn} \\
CCAligned & \texttt{qd} & \texttt{kac} \\
CCAligned & \texttt{cx} & \texttt{ceb} \\
\midrule
mC4 & \texttt{iw} & \texttt{he}  \\
\midrule
OSCAR & \texttt{eml} & \texttt{egl}  \\
OSCAR & \texttt{als} & \texttt{gsw}  \\
OSCAR & \texttt{sh} & \texttt{hbs}  \\
\midrule
WikiMatrix & \texttt{sh} & \texttt{hbs}  \\
\bottomrule
\end{tabular}
\caption{Miscellaneous errors in language codes.} 
\label{tab:misc_codes}
\end{table}

\section{Complete Error Taxonomy and Instructions}~\label{app:taxonomy}
In addition to the examples given in Table \ref{tab:examples}, raters were provided with the following verbal notes on the error codes:
\begin{itemize}
    \item \textbf{\texttt{CC}: Correct translation, natural sentence:} It's OK if it's a sentence fragment instead of a whole sentence, as long as it is not too short (about 5 words or greater). The translation does not have to be perfect.
    \item \textbf{\texttt{\texttt{CS}}: Correct Translation, but single word or short phrase:} Also includes highly repeated short phrases, like ``the cat the cat the cat the cat the cat ..."
    \item \textbf{\texttt{CB}: Correct translation, but boilerplate: } This can be auto-generated or formulaic content, or content that one deems ``technically correct but generally not very useful to NLP models". Unfortunately, it's often not clear what should be counted as boilerplate...do your best.
    \item \textbf{\texttt{X}: Incorrect translation} [for parallel sentences] both source and target are in the correct language, but they are not adequate translations.
    \item \textbf{\texttt{WL}: Wrong language} For short sentences, especially with proper nouns, there is often a fine line between ``Wrong language" and ``Not language". Do your best. 
    \item \textbf{\texttt{NL}: Not language} At least one of source and target are not linguistic content. Any sentence consisting only of a proper noun (e.g. ``Tyrone Ping") should be marked as \texttt{NL}.
    \item \textbf{\texttt{U}: Unknown} for sentences that need verification by a native speaker. This is an auxiliary label that is resolved in most cases.
\end{itemize}

\section{Methodological Notes}\label{app:strategies}

A surprising amount of work can be done without being an expert in the languages involved. The easiest approach is simply to search the internet for the sentence, which usually results in finding the exact page the sentence came from, which in turn frequently contains clues like language codes in the URL, or a headline like \textit{News in X language}, sometimes with references to a translated version of the same page. However, for the cases where this is insufficient, here are a few tips, tricks, and observations.

\paragraph{No Skills Required:}
Things that do not require knowledge of the language(s) in question.
\begin{enumerate}
    \item ``Not language'' can usually be identified by anyone who can read the script, though there are tricky cases with proper nouns.
    \item Frequently, ``parallel" sentences contain different numbers in the source and target (especially autogenerated content), and are easy to disqualify.
    \item Errors tend to repeat. If a word is mistranslated once, it will often be mistranslated many more times throughout a corpus, making it easy to spot.
\end{enumerate}

\paragraph{Basic Research Required:}
Things that do not require knowledge of the language(s) in question but can be done with basic research.
\begin{enumerate}
    \item If it's written in the wrong script it's considered wrong language. (Sometimes the writing system is indicated in the published corpus, e.g. \texttt{bg-Latn}, but usually the language has a ``default" script defined by ISO.)
    \item Some types of texts come with inherent labels or markers, such as enumerators or verse numbers.
    %For example, much of CCAligned's Odia text is Christian Bible verses, which are preceded by an identifier like ``Matt 12:37". 
    \item When all else fails, search the internet for the whole sentence or n-grams thereof! If the whole sentence can be found, frequently the language is betrayed by the web page (the language's autonym is useful in this case).
\end{enumerate}

\section{Complete Audit Results}\label{app:stats}
Tables \ref{tab:ccaligned-full}, \ref{tab:wikimatrix-full}, \ref{tab:paracrawl-full}, \ref{tab:mc4-full} and \ref{tab:oscar-full} give the complete annotation percentages for CCAligned, WikiMatrix, ParaCrawl, mC4 and OSCAR, respectively. For each annotation label, we report the ratio of the annotated sentences (of max 100 sentences) that were assigned that label by the primary annotator. Repeated annotations done for agreement measurement are not included. The \texttt{C} column aggregates all correct sub-codes (\texttt{CC}, \texttt{CS}, \texttt{CB}). We also report the total number of sentences that each dataset contains for each language and the average sentence length for the audited sentences to illustrate differences across languages. The original language codes as they are published with the datasets are maintained for the sake of consistency (but should be handled with care in future work, see Section~\ref{sec:codes}), and those with less than 20\% correct sentences are highlighted.

\input{full_tables}

\end{document}

%% file: full_tables.tex
%%% CCALIGNED %%%

\begin{table*}[hbt!]
\centering\small
\resizebox*{0.9\textwidth}{!}{ %\textheight}{%
\begin{tabular}{l|rrrr|rrrr|rr}
\toprule
{} &       C &           CC &     CS &     CB & X &      WL &      NL &     porn & \#sentences & avg target length\\
\midrule
\textbf{en-sz\_PL}     &  0.00\% &  0.00\% &  0.00\% &  0.00\% &  0.00\% &  8.33\% & 91.67\% &  0.00\% &       12 &  71.42 \\
\textbf{en-mt\_MT}     &  3.85\% &  0.00\% &  3.85\% &  0.00\% & 50.00\% & 26.92\% & 19.23\% &  0.00\% &       26 &  12.58 \\
\textbf{en-tz\_MA}     & 12.12\% &  6.06\% &  6.06\% &  0.00\% & 45.45\% & 36.36\% &  6.06\% &  0.00\% &       33 &  57.33 \\
\textbf{en-zz\_TR}     &  0.00\% &  0.00\% &  0.00\% &  0.00\% &  8.82\% & 61.76\% & 29.41\% &  0.00\% &       34 &  46.53 \\
\textbf{en-kg\_AO}     &  1.35\% &  0.00\% &  1.35\% &  0.00\% & 14.86\% &  2.70\% & 81.08\% &  0.00\% &       74 &  29.20 \\
\textbf{en-qa\_MM}     & 11.03\% &  5.88\% &  3.68\% &  1.47\% & 72.06\% &  3.68\% & 13.24\% &  0.00\% &      136 &  55.28 \\
\textbf{en-bm\_ML}     &  6.04\% &  4.03\% &  2.01\% &  0.00\% & 26.85\% &  6.71\% & 60.40\% &  0.00\% &      149 &  32.19 \\
\textbf{en-az\_IR}     &  6.93\% &  6.93\% &  0.00\% &  0.00\% & 20.79\% & 13.86\% & 58.42\% &  0.00\% &      158 & 115.85 \\
\textbf{en-qd\_MM}     &  7.92\% &  4.95\% &  1.98\% &  0.99\% & 81.19\% &  3.96\% &  6.93\% &  0.00\% &      179 &  60.34 \\
en-ay\_BO     & 51.00\% & 33.00\% & 18.00\% &  0.00\% & 29.00\% &  3.00\% & 17.00\% &  0.00\% &      475 &  92.19 \\
\textbf{en-ak\_GH}     & 14.23\% & 13.60\% &  0.63\% &  0.00\% & 46.86\% & 19.25\% & 19.67\% &  0.00\% &      478 &  45.85 \\
en-st\_ZA     & 48.57\% & 42.14\% &  0.00\% &  6.43\% & 40.71\% &  1.43\% &  9.29\% &  0.00\% &      904 & 111.83 \\
en-ve\_ZA     & 60.40\% & 29.70\% & 21.78\% &  8.91\% & 28.71\% &  3.96\% &  6.93\% &  0.00\% &     1555 &  82.99 \\
en-ts\_ZA     & 51.49\% & 34.65\% & 11.88\% &  4.95\% & 40.59\% &  2.97\% &  4.95\% &  0.00\% &     1967 &  73.93 \\
en-or\_IN     & 42.61\% &  6.09\% & 24.35\% & 12.17\% & 38.26\% &  9.57\% &  9.57\% &  0.00\% &     5526 &  71.39 \\
\textbf{en-ns\_ZA }    &  4.00\% &  2.00\% &  0.00\% &  2.00\% & 23.00\% & 15.00\% & 58.00\% &  4.00\% &    14138 &  33.52 \\
\textbf{en-lg\_UG}     &  6.00\% &  0.00\% &  6.00\% &  0.00\% & 68.00\% & 17.00\% &  9.00\% &  2.00\% &    14701 &  15.83 \\
\textbf{en-ln\_CD}     &  8.00\% &  4.00\% &  3.00\% &  1.00\% & 14.00\% &  4.00\% & 74.00\% &  4.00\% &    21562 &  28.80 \\
\textbf{en-om\_KE}     &  2.00\% &  2.00\% &  0.00\% &  0.00\% & 31.00\% & 38.00\% & 29.00\% & 24.00\% &    22206 &  23.83 \\
\textbf{en-ss\_SZ}     & 12.65\% &  9.04\% &  3.61\% &  0.00\% & 13.25\% & 24.10\% & 50.00\% & 13.86\% &    22960 &  25.30 \\
\textbf{en-te\_IN\_rom} &  0.00\% &  0.00\% &  0.00\% &  0.00\% & 25.00\% &  8.00\% & 67.00\% &  5.00\% &    25272 &  24.21 \\
\textbf{en-cb\_IQ}     &  4.00\% &  1.00\% &  3.00\% &  0.00\% & 30.00\% & 18.00\% & 48.00\% & 11.00\% &    52297 &  30.04 \\
\textbf{en-tn\_BW}     &  0.00\% &  0.00\% &  0.00\% &  0.00\% &  6.90\% &  8.97\% & 63.45\% & 10.34\% &    71253 &  16.80 \\
\textbf{en-ff\_NG}     &  0.00\% &  0.00\% &  0.00\% &  0.00\% &  0.00\% &  8.00\% & 92.00\% &  2.00\% &    73022 &  33.59 \\
\textbf{en-sn\_ZW}     &  5.00\% &  1.00\% &  3.00\% &  1.00\% & 81.00\% & 14.00\% &  0.00\% &  0.00\% &    86868 & 102.59 \\
\textbf{en-wo\_SN}     &  0.00\% &  0.00\% &  0.00\% &  0.00\% &  1.71\% &  3.31\% & 94.98\% & 18.46\% &    88441 &  27.25 \\
\textbf{en-br\_FR}     & 17.00\% &  3.00\% &  1.00\% & 13.00\% & 37.00\% & 14.00\% & 32.00\% &  1.00\% &   115128 &  41.68 \\
en-zu\_ZA     & 55.00\% & 39.00\% &  3.00\% & 13.00\% & 30.00\% &  7.00\% &  8.00\% &  3.00\% &   126101 &  79.32 \\
en-ku\_TR     & 36.52\% & 12.17\% & 13.04\% & 11.30\% & 33.04\% & 28.70\% &  1.74\% &  1.74\% &   137874 &  90.51 \\
en-ig\_NG     & 58.00\% & 49.00\% &  3.00\% &  6.00\% & 29.00\% & 12.00\% &  1.00\% &  0.00\% &   148146 &  83.42 \\
en-kn\_IN     & 46.00\% &  9.00\% &  6.00\% & 31.00\% & 46.00\% &  2.00\% &  5.00\% &  4.00\% &   163921 &  70.20 \\
en-yo\_NG     & 34.93\% &  6.16\% & 10.96\% & 17.81\% & 34.93\% & 12.33\% & 17.81\% &  0.00\% &   175192 &  75.01 \\
en-ky\_KG     & 44.12\% & 24.51\% & 17.65\% &  1.96\% & 33.33\% & 22.55\% &  0.00\% &  0.98\% &   240657 &  69.56 \\
en-tg\_TJ     & 46.08\% & 18.63\% & 24.51\% &  2.94\% & 32.35\% & 20.59\% &  0.98\% &  4.90\% &   251865 &  75.31 \\
en-ha\_NG     & 30.00\% & 25.00\% &  3.00\% &  2.00\% & 49.00\% &  9.00\% & 12.00\% &  1.00\% &   339176 &  60.78 \\
en-am\_ET     & 59.11\% & 35.47\% &  2.46\% & 21.18\% & 37.44\% &  2.96\% &  0.49\% &  0.00\% &   346517 &  58.29 \\
en-km\_KH     & 56.12\% & 12.24\% & 33.67\% & 10.20\% & 42.86\% &  1.02\% &  0.00\% &  0.00\% &   412381 &  71.35 \\
en-ne\_NP     & 47.00\% & 10.00\% & 13.00\% & 24.00\% & 15.00\% &  8.00\% & 30.00\% & 14.00\% &   487155 &  79.14 \\
en-su\_ID     & 35.00\% & 15.00\% & 15.00\% &  5.00\% & 13.00\% & 13.00\% & 39.00\% &  0.00\% &   494142 &  57.08 \\
\textbf{en-ur\_PK\_rom} &  0.50\% &  0.00\% &  0.50\% &  0.00\% & 18.91\% & 27.36\% & 53.23\% &  5.47\% &   513123 &  18.41 \\
en-ht\_HT     & 55.67\% &  8.25\% & 10.31\% & 37.11\% & 35.05\% &  6.19\% &  3.09\% &  1.03\% &   558167 & 101.95 \\
en-mn\_MN     & 33.00\% &  8.00\% & 14.00\% & 11.00\% & 42.00\% &  7.00\% & 18.00\% & 12.00\% &   566885 &  44.43 \\
en-te\_IN     & 69.00\% & 42.00\% & 11.00\% & 16.00\% & 27.00\% &  1.00\% &  3.00\% &  1.00\% &   581651 &  97.95 \\
en-kk\_KZ     & 68.32\% & 40.59\% & 18.81\% &  8.91\% & 18.81\% &  8.91\% &  3.96\% &  1.98\% &   689651 &  72.36 \\
en-be\_BY     & 90.00\% & 57.00\% & 13.00\% & 20.00\% & 10.00\% &  0.00\% &  0.00\% &  2.00\% &  1125772 & 118.45 \\
en-af\_ZA     & 63.00\% & 40.00\% & 23.00\% &  0.00\% & 31.00\% &  2.00\% &  4.00\% & 12.00\% &  1504061 & 105.45 \\
\textbf{en-jv\_ID}     &  5.05\% &  1.01\% &  1.01\% &  3.03\% & 25.25\% & 10.10\% & 59.60\% &  8.08\% &  1513974 &  18.34 \\

\textbf{en-hi\_IN\_rom} &  1.00\% &  0.00\% &  0.00\% &  1.00\% & 39.00\% & 21.00\% & 39.00\% &  8.00\% &  3789571 &  18.13 \\
en-lv\_LV     & 59.00\% & 37.00\% &  9.00\% & 13.00\% & 31.00\% &  7.00\% &  3.00\% & 14.00\% &  4850957 &  83.67 \\
\textbf{en-ar\_AR\_rom} &  0.00\% &  0.00\% &  0.00\% &  0.00\% &  0.00\% &  4.00\% & 96.00\% &  4.00\% &  5584724 &  16.69 \\
\textbf{en-tl\_XX}     & 13.00\% &  6.00\% &  3.00\% &  4.00\% & 24.00\% & 26.00\% & 37.00\% &  5.00\% &  6593250 &  37.03 \\
en-uk\_UA     & 63.00\% & 42.00\% &  8.00\% & 13.00\% & 35.00\% &  1.00\% &  1.00\% &  5.00\% &  8547348 &  67.88 \\
en-zh\_TW     & 46.00\% & 11.00\% & 31.00\% &  4.00\% & 47.00\% &  6.00\% &  1.00\% &  1.00\% &  8778971 &  24.89 \\
en-el\_GR     & 49.00\% & 15.00\% &  5.00\% & 29.00\% & 38.00\% &  3.00\% & 10.00\% &  8.00\% &  8878492 &  54.90 \\
en-nl\_NL     & 46.00\% & 27.00\% & 19.00\% &  0.00\% & 49.00\% &  2.00\% &  3.00\% &  0.00\% &  36324231  &  85.95 \\
en-da\_DK     & 54.00\% & 31.00\% & 18.00\% &  5.00\% & 29.00\% &  5.00\% & 12.00\% &  7.00\% & 10738582 &  73.99 \\
en-vi\_VN     & 31.00\% & 18.00\% &  0.00\% & 13.00\% & 54.00\% &  1.00\% & 14.00\% &  6.00\% & 12394379 &  74.19 \\
en-sv\_SE     & 97.00\% & 91.00\% &  3.00\% &  3.00\% &  0.00\% &  3.00\% &  0.00\% &  0.00\% & 12544075 & 103.91 \\
en-zh\_CN     & 57.29\% & 22.92\% & 12.50\% & 21.88\% & 31.25\% &  1.04\% & 10.42\% &  1.04\% & 15181410 &  33.55 \\
en-tr\_TR     & 45.00\% & 14.50\% & 14.00\% & 16.50\% & 44.50\% &  5.00\% &  5.50\% &  4.00\% & 20282339 &  83.80 \\
en-ja\_XX     & 57.00\% & 35.00\% & 21.00\% &  1.00\% & 34.00\% &  6.00\% &  0.00\% &  0.00\% & 26201214 &  34.44 \\
en-pt\_XX     & 66.34\% & 36.63\% & 10.89\% & 18.81\% & 20.79\% &  3.96\% &  8.91\% &  0.00\% & 46525410 &  87.20 \\
en-it\_IT     & 36.00\% & 14.00\% & 18.00\% &  4.00\% & 60.00\% &  1.00\% &  3.00\% &  0.00\% & 58022366 &  97.44 \\
en-de\_DE     & 62.00\% & 29.00\% & 14.00\% & 19.00\% & 28.00\% &  2.00\% &  8.00\% &  2.00\% & 92597196 &  78.08 \\
en-es\_XX     & 58.42\% & 16.83\% & 25.74\% & 15.84\% & 22.77\% &  2.97\% & 15.84\% &  4.95\% & 98351611 &  72.18 \\
%\midrule
%\textit{mean} & 27.01\% & 29.35\% & 8.62\% & 28.97\% & 14.48\% & 6.49\% & 5.89\% &     0.00\% & 5.26\% & \\
%nl  36324231 after id_ID \\
\bottomrule
\end{tabular}%
}
\caption{Audit results for a sample of 100 sentences from \textbf{CCAligned} for each language pair, compared to the number of sentences available in the dataset. If fewer than 100 sentences were available, all sentences were audited. Language codes are as originally published.  The length is measured in number of characters and averaged across the audited portion of each corpus. Languages with less than 20\% correct sentences are boldfaced.}

\label{tab:ccaligned-full}
\end{table*}

% template frame:
%\begin{table*}
%\centering
%\resizebox*{0.8\textwidth}{\textheight}{%
% [INSERT TABLE]
%}
%\caption{Audit results for a sample of 100 sentences from CCAligned for each language.}
%\end{table*}

\clearpage

%%% WIKIMATRIX %%%
\begin{table*}[hbt!]
\centering
\resizebox{0.9\textwidth}{!}{%
\begin{tabular}{l|rrrr|rrrr|rr}
\toprule
{} &      C &    CC &    CS &     CB &   X &     WL &    NL &       porn & \# sentences & avg target length\\
\midrule
\textbf{en-ug}     & 12.87\% &  8.91\% & 1.98\% &  1.98\% & 72.28\% &  9.90\% & 1.98\% & 0.00\% &   22012 &  95.55 \\
en-mwl    & 27.00\% & 26.00\% & 0.00\% &  1.00\% & 73.00\% &  0.00\% & 0.00\% & 0.00\% &   33899 & 135.26 \\
\textbf{en-tg}     &  0.00\% &  0.00\% & 0.00\% &  0.00\% & 95.10\% &  3.92\% & 0.98\% & 0.00\% &   37975 &  88.87 \\
\textbf{en-ne}     & 13.00\% &  7.00\% & 6.00\% &  0.00\% & 60.00\% & 23.00\% & 4.00\% & 0.00\% &   40549 &  69.26 \\
\textbf{en-ka}     & 11.88\% &  2.97\% & 2.97\% &  5.94\% & 73.27\% & 10.89\% & 2.97\% & 0.00\% &   41638 & 144.74 \\
\textbf{en-lmo }   & 12.75\% & 11.76\% & 0.00\% &  0.98\% & 81.37\% &  4.90\% & 0.98\% & 0.00\% &   43790 &  89.38 \\
en-io     & 28.00\% & 27.00\% & 0.00\% &  1.00\% & 69.00\% &  2.00\% & 1.00\% & 0.00\% &   45999 &  83.26 \\
\textbf{en-jv}     & 13.73\% &  9.80\% & 0.00\% &  3.92\% & 70.59\% & 12.75\% & 2.94\% & 0.00\% &   48301 &  91.87 \\
en-wuu    & 23.23\% & 14.14\% & 7.07\% &  2.02\% & 65.66\% &  7.07\% & 4.04\% & 0.00\% &   51024 &  34.77 \\
\textbf{br-en}     &  8.70\% &  7.61\% & 1.09\% &  0.00\% & 82.61\% &  4.35\% & 0.00\% & 0.00\% &   58400 &  90.68 \\
\textbf{bar-en}    &  6.00\% &  6.00\% & 0.00\% &  0.00\% & 75.00\% & 16.00\% & 3.00\% & 0.00\% &   67394 & 103.51 \\
\textbf{en-kk}     &  5.00\% &  2.00\% & 2.00\% &  1.00\% & 81.00\% & 14.00\% & 0.00\% & 0.00\% &  109074 &  56.03 \\
en-sw     & 33.33\% & 27.27\% & 4.04\% &  2.02\% & 64.65\% &  2.02\% & 0.00\% & 0.00\% &  138590 & 111.61 \\
\textbf{en-nds}    &  1.96\% &  1.96\% & 0.00\% &  0.00\% & 95.10\% &  1.96\% & 0.98\% & 0.00\% &  178533 &  91.95 \\
be-en     & 26.00\% & 24.00\% & 2.00\% &  0.00\% & 73.00\% &  1.00\% & 0.00\% & 0.00\% &  257946 & 121.22 \\
en-hi     & 36.27\% & 32.35\% & 0.98\% &  2.94\% & 59.80\% &  0.98\% & 2.94\% & 0.00\% &  696125 &  96.77 \\
en-ko     & 48.04\% & 33.33\% & 2.94\% & 11.76\% & 48.04\% &  2.94\% & 0.98\% & 0.00\% & 1345630 &  55.18 \\
en-uk     & 87.00\% & 84.00\% & 2.00\% &  1.00\% & 10.00\% &  1.00\% & 2.00\% & 0.00\% & 2576425 & 104.39 \\
en-it     & 42.00\% & 42.00\% & 0.00\% &  0.00\% & 58.00\% &  0.00\% & 0.00\% & 0.00\% & 4626048 & 140.27 \\
en-simple & 37.62\% & 24.75\% & 0.00\% & 12.87\% & 56.44\% &  2.97\% & 2.97\% & 0.00\% &     N/A &  77.53 \\
\bottomrule
\end{tabular}%
}
\caption{Audit results for a sample of 100 sentences from \textbf{WikiMatrix} for each language pair, compared to the number of sentences available in the dataset. Language codes are as originally published. The length is measured in number of characters and averaged across the audited portion of each corpus. Languages with less than 20\% correct sentences are boldfaced.}

\label{tab:wikimatrix-full}
\end{table*}

%%% PARACRAWL %%%
\begin{table*}[hbt!]
\centering
\resizebox{0.9\textwidth}{!}{%
\begin{tabular}{l|rrrr|rrrr|rr}
\toprule
{} &      C &      CC &     CS &     CB &     X &    WL &    NL &  porn & \# sentences & avg target length\\
\midrule
en-so & 80.81\% & 61.62\% &  1.01\% & 18.18\% & 14.14\% &  5.05\% &  0.00\% & 0.00\% &     14879 & 189.83 \\
en-ps & 72.00\% & 53.00\% &  9.00\% & 10.00\% & 17.00\% & 10.00\% &  0.00\% & 0.00\% &     26321 & 141.01 \\
en-my & 45.00\% &  9.00\% & 16.00\% & 20.00\% & 32.00\% &  9.00\% & 14.00\% & 0.00\% &     31374 & 147.07 \\
en-km & 76.00\% & 51.00\% & 13.00\% & 12.00\% & 18.00\% &  6.00\% &  0.00\% & 0.00\% &     65113 & 121.20 \\
en-ne & 73.00\% & 48.00\% &  1.00\% & 24.00\% & 23.00\% &  2.00\% &  0.00\% & 0.00\% &     92084 & 153.42 \\
en-sw & 85.00\% & 60.00\% & 15.00\% & 10.00\% & 11.00\% &  2.00\% &  2.00\% & 0.00\% &    132517 & 167.34 \\
en-si & 37.00\% & 31.00\% &  6.00\% &  0.00\% & 62.00\% &  0.00\% &  1.00\% & 0.00\% &    217407 & 123.06 \\
en-nn & 35.92\% & 24.27\% &  8.74\% &  2.91\% & 49.51\% & 13.59\% &  0.97\% & 0.00\% &    323519 &  56.24 \\
es-eu & 88.00\% & 66.00\% & 15.00\% &  7.00\% & 10.00\% &  1.00\% &  1.00\% & 0.00\% &    514610 & 121.31 \\
es-gl & 89.00\% & 46.00\% &  6.00\% & 37.00\% &  4.00\% &  7.00\% &  0.00\% & 0.00\% &   1222837 & 107.88 \\
en-ru & 81.00\% & 73.00\% &  6.00\% &  2.00\% & 19.00\% &  0.00\% &  0.00\% & 6.00\% &   5377911 & 101.28 \\
en-bg & 95.15\% & 85.44\% &  0.97\% &  8.74\% &  4.85\% &  0.00\% &  0.00\% & 0.97\% &   6470710 & 112.29 \\
es-ca & 80.00\% & 54.00\% & 19.00\% &  7.00\% & 11.00\% &  9.00\% &  0.00\% & 5.00\% &   6870183 & 107.21 \\
en-el & 91.59\% & 68.22\% &  0.93\% & 22.43\% &  7.48\% &  0.93\% &  0.00\% & 0.00\% &   9402646 & 135.66 \\
en-pl & 94.12\% & 76.47\% &  0.98\% & 16.67\% &  3.92\% &  1.96\% &  0.00\% & 0.98\% &  13744860 &  95.95 \\
en-nl & 49.00\% & 32.00\% & 17.00\% &  0.00\% & 46.00\% &  3.00\% &  2.00\% & 0.00\% &  31295016 &  95.05 \\
en-pt & 93.07\% & 92.08\% &  0.00\% &  0.99\% &  4.95\% &  1.98\% &  0.00\% & 0.00\% &  31486963 & 108.68 \\
en-it & 60.82\% & 36.08\% & 16.49\% &  8.25\% & 38.14\% &  0.00\% &  1.03\% & 0.00\% &  40798278 & 127.55 \\
en-es & 87.00\% & 54.00\% & 20.00\% & 13.00\% & 12.00\% &  0.00\% &  1.00\% & 0.50\% &  78662122 & 119.72 \\
en-de & 82.83\% & 64.65\% & 13.13\% &  5.05\% & 13.13\% &  3.03\% &  1.01\% & 0.00\% &  82638202 & 111.43 \\
en-fr & 89.62\% & 82.08\% &  4.72\% &  2.83\% & 10.38\% &  0.00\% &  0.00\% & 0.00\% & 104351522 & 144.20 \\
\bottomrule
\end{tabular} %
}
\caption{Audit results for a sample of 100 sentences from \textbf{ParaCrawl} for each language pair, compared to the number of sentences available in the dataset. Language codes are as originally published.  The length is measured in number of characters and averaged across the audited portion of each corpus.}
\label{tab:paracrawl-full}
\end{table*}

\clearpage

%%% mC4 %%%

\begin{table*}[hbt!]
\centering
\resizebox*{0.9\textwidth}{!}{%
\begin{tabular}{l|rrrr|rrr|rr}
\toprule
{} &      C &    CC &    CS &     CB &     WL &    NL &       porn & \# sentences & avg length\\
\midrule
yo      & 84.69\% & 71.43\% &  2.04\% & 11.22\% & 14.29\% &  1.02\% & 0.00\% &      46214 & 117.71 \\
st      & 56.70\% & 42.27\% & 14.43\% &  0.00\% & 35.05\% &  8.25\% & 0.00\% &      66837 & 132.13 \\
haw     & 44.90\% & 34.69\% &  1.02\% &  9.18\% & 33.67\% & 21.43\% & 1.02\% &      84312 & 129.99 \\
ig      & 55.91\% & 41.73\% & 10.24\% &  3.94\% &  0.00\% & 44.09\% & 0.79\% &      92909 &  98.03 \\
sm      & 60.20\% & 58.16\% &  2.04\% &  0.00\% & 27.55\% & 12.24\% & 0.00\% &      98467 & 126.42 \\
ha      & 80.81\% & 79.80\% &  1.01\% &  0.00\% & 14.14\% &  5.05\% & 2.02\% &     247479 & 155.76 \\
su      & 59.60\% & 58.59\% &  1.01\% &  0.00\% & 25.25\% & 15.15\% & 2.02\% &     280719 & 107.10 \\
sn      & 36.63\% & 32.67\% &  2.97\% &  0.99\% & 58.42\% &  4.95\% & 0.00\% &     326392 & 145.59 \\
mg      & 57.00\% & 57.00\% &  0.00\% &  0.00\% & 18.00\% & 25.00\% & 0.00\% &     345040 & 116.23 \\
pa      & 78.30\% & 68.87\% &  3.77\% &  5.66\% &  4.72\% & 10.38\% & 0.00\% &     363399 & 134.43 \\
ga      & 76.77\% & 58.59\% &  6.06\% & 12.12\% & 10.10\% & 13.13\% & 0.00\% &     465670 & 147.35 \\
co      & 33.00\% & 29.00\% &  2.00\% &  2.00\% & 48.00\% & 19.00\% & 0.00\% &     494913 & 195.30 \\
zu      & 51.00\% & 48.00\% &  2.00\% &  1.00\% & 30.00\% & 19.00\% & 0.00\% &     555458 & 137.81 \\
jv      & 52.73\% & 19.09\% & 19.09\% & 14.55\% & 40.00\% &  7.27\% & 1.82\% &     581528 &  97.96 \\
km      & 92.86\% & 92.86\% &  0.00\% &  0.00\% &  7.14\% &  0.00\% & 0.00\% &     756612 & 162.57 \\
kn      & 85.15\% & 73.27\% &  3.96\% &  7.92\% &  2.97\% &  9.90\% & 0.00\% &    1056849 & 105.39 \\
fy      & 56.73\% & 50.00\% &  3.85\% &  2.88\% & 39.42\% &  3.85\% & 0.00\% &    1104359 & 234.25 \\
te      & 89.00\% & 76.00\% &  9.00\% &  4.00\% &  3.00\% &  8.00\% & 0.00\% &    1188243 & 108.49 \\
la      & 82.31\% & 65.38\% &  6.15\% & 10.77\% & 10.00\% &  7.69\% & 0.00\% &    1674463 &  67.25 \\
be      & 92.04\% & 86.73\% &  2.65\% &  2.65\% &  4.42\% &  3.54\% & 0.00\% &    1742030 & 110.86 \\
af      & 76.00\% & 76.00\% &  0.00\% &  0.00\% & 15.00\% &  9.00\% & 0.00\% &    2152243 &  99.52 \\
\textbf{lb}      & 17.48\% & 17.48\% &  0.00\% &  0.00\% &  7.77\% & 74.76\% & 0.00\% &    2740336 & 481.68 \\
ne      & 78.35\% & 77.32\% &  1.03\% &  0.00\% & 21.65\% &  0.00\% & 0.00\% &    2942785 & 102.88 \\
sr      & 93.69\% & 85.59\% &  7.21\% &  0.90\% &  5.41\% &  0.00\% & 0.00\% &    3398483 & 131.72 \\
gl      & 67.62\% & 57.14\% & 10.48\% &  0.00\% & 13.33\% & 17.14\% & 0.00\% &    4549465 & 151.45 \\
bn      & 93.00\% & 86.00\% &  1.00\% &  6.00\% &  3.00\% &  4.00\% & 0.00\% &    7444098 &  92.60 \\
mr      & 40.00\% & 35.24\% &  2.86\% &  1.90\% & 49.52\% & 10.48\% & 0.00\% &    7774331 & 281.94 \\
sl      & 92.08\% & 82.18\% &  4.95\% &  4.95\% &  2.97\% &  4.95\% & 0.00\% &    8499456 & 149.45 \\
hi      & 80.30\% & 76.77\% &  1.01\% &  2.53\% & 19.70\% &  0.00\% & 2.53\% &   18507273 & 105.54 \\
bg      & 80.90\% & 75.88\% &  2.51\% &  2.51\% &  2.01\% & 17.09\% & 0.00\% &   23409799 &  93.86 \\
uk      & 95.48\% & 81.41\% &  7.54\% &  6.53\% &  2.01\% &  2.51\% & 0.00\% &   38556465 & 116.79 \\
ro      & 94.95\% & 78.79\% & 12.12\% &  4.04\% &  3.03\% &  2.02\% & 0.00\% &   45738857 & 130.08 \\
sv      & 91.18\% & 84.31\% &  2.94\% &  3.92\% &  4.90\% &  3.92\% & 1.96\% &   48570979 & 114.45 \\
zh      & 92.00\% & 87.00\% &  1.00\% &  4.00\% &  1.00\% &  7.00\% & 0.00\% &   54542308 &  94.77 \\
ja      & 99.00\% & 89.00\% &  6.00\% &  4.00\% &  0.00\% &  1.00\% & 1.00\% &   87337884 &  59.94 \\
tr      & 95.96\% & 88.89\% &  0.00\% &  7.07\% &  3.54\% &  0.51\% & 0.00\% &   87595290 & 152.75 \\
nl      & 92.08\% & 85.15\% &  6.93\% &  0.00\% &  1.98\% &  5.94\% & 0.00\% &   96210458 & 103.67 \\
pl      & 96.00\% & 82.00\% &  7.00\% &  7.00\% &  2.00\% &  2.00\% & 0.00\% &  126164277 & 170.70 \\
pt      & 86.00\% & 79.00\% &  4.00\% &  3.00\% &  2.00\% & 12.00\% & 1.00\% &  169239084 & 133.51 \\
it      & 92.00\% & 79.00\% &  9.00\% &  4.00\% &  1.00\% &  7.00\% & 0.00\% &  186404508 & 180.26 \\
fr      & 92.00\% & 82.00\% &  7.00\% &  3.00\% &  1.00\% &  7.00\% & 0.00\% &  332674575 & 143.69 \\
de      & 91.18\% & 77.45\% &  7.84\% &  5.88\% &  6.86\% &  1.96\% & 0.00\% &  397006993 & 107.71 \\
ru      & 91.06\% & 69.11\% & 11.38\% & 10.57\% &  4.07\% &  4.88\% & 0.00\% &  755585265 & 109.28 \\
en      & 93.94\% & 83.84\% &  8.08\% &  2.02\% &  1.01\% &  5.05\% & 0.00\% & 3079081989 & 130.97 \\
\textbf{bg\_latn} &  9.09\% &  9.09\% &  0.00\% &  0.00\% & 51.52\% & 39.39\% & 1.01\% &        N/A & 139.92 \\
\textbf{ja\_latn} & 13.00\% &  7.00\% &  4.00\% &  2.00\% & 60.00\% & 27.00\% & 0.00\% &        N/A & 218.92 \\
ru\_latn & 36.45\% & 25.23\% & 10.28\% &  0.93\% & 34.58\% & 28.97\% & 0.93\% &        N/A & 123.14 \\
\textbf{zh\_latn} &  5.00\% &  4.00\% &  1.00\% &  0.00\% & 64.00\% & 31.00\% & 0.00\% &        N/A & 186.84 \\
\bottomrule
\end{tabular}%
}
\caption{Audit results for a sample of 100 sentences from \textbf{mC4} for each language, compared to the number of sentences available in the dataset. Language codes are as originally published. The length is measured in number of characters and averaged across the audited portion of each corpus. Languages with less than 20\% correct sentences are boldfaced.}
\label{tab:mc4-full}
\end{table*}
\clearpage

%%% OSCAR %%%
\begin{table*}[hbt!]
\centering
\resizebox*{0.9\textwidth}{!}{%
\begin{tabular}{l|rrrr|rrr|rr}
\toprule
{} &      C &    CC &    CS &     CB &     WL &    NL &       porn & \# sentences & avg length\\
\midrule
diq & 100.00\% & 100.00\% & 0.00\% &  0.00\% &   0.00\% &   0.00\% & 0.00\% &          1 & 131.00 \\
\textbf{bcl} &   0.00\% &   0.00\% & 0.00\% &  0.00\% &   0.00\% & 100.00\% & 0.00\% &          1 & 623.00 \\
\textbf{cbk} &   0.00\% &   0.00\% & 0.00\% &  0.00\% & 100.00\% &   0.00\% & 0.00\% &          1 & 519.00 \\
pam & 100.00\% & 100.00\% & 0.00\% &  0.00\% &   0.00\% &   0.00\% & 0.00\% &          2 & 139.00 \\
bar &  25.00\% &  25.00\% & 0.00\% &  0.00\% &   0.00\% &  75.00\% & 0.00\% &          4 &  53.50 \\
myv & 100.00\% & 100.00\% & 0.00\% &  0.00\% &   0.00\% &   0.00\% & 0.00\% &          5 & 127.00 \\
\textbf{yue} &   0.00\% &   0.00\% & 0.00\% &  0.00\% &  57.14\% &  42.86\% & 0.00\% &          7 & 177.00 \\
mwl &  57.14\% &  57.14\% & 0.00\% &  0.00\% &  42.86\% &   0.00\% & 0.00\% &          7 & 141.00 \\
\textbf{frr} &   0.00\% &   0.00\% & 0.00\% &  0.00\% &   0.00\% & 100.00\% & 0.00\% &          9 & 231.56 \\
ht  &  30.00\% &  30.00\% & 0.00\% &  0.00\% &   0.00\% &  70.00\% & 0.00\% &         10 & 329.10 \\
ie  &  30.00\% &  30.00\% & 0.00\% &  0.00\% &  30.00\% &  40.00\% & 0.00\% &         11 & 121.70 \\
scn & 100.00\% & 100.00\% & 0.00\% &  0.00\% &   0.00\% &   0.00\% & 0.00\% &         17 & 155.59 \\
tyv &  96.15\% &  96.15\% & 0.00\% &  0.00\% &   0.00\% &   3.85\% & 0.00\% &         26 & 167.96 \\
mai &  79.31\% &  75.86\% & 0.00\% &  3.45\% &  20.69\% &   0.00\% & 0.00\% &         29 & 141.17 \\
bxr & 100.00\% & 100.00\% & 0.00\% &  0.00\% &   0.00\% &   0.00\% & 0.00\% &         37 & 160.76 \\
dsb & 100.00\% &  97.56\% & 0.00\% &  2.44\% &   0.00\% &   0.00\% & 0.00\% &         41 & 155.15 \\
\textbf{so}  &   0.00\% &   0.00\% & 0.00\% &  0.00\% &  28.57\% &  71.43\% & 0.00\% &         42 & 208.24 \\
rm  & 100.00\% & 100.00\% & 0.00\% &  0.00\% &   0.00\% &   0.00\% & 0.00\% &         47 & 137.66 \\
nah & 100.00\% &  96.67\% & 0.00\% &  3.33\% &   0.00\% &   0.00\% & 0.00\% &         60 & 164.53 \\
\textbf{nap} &   0.00\% &   0.00\% & 0.00\% &  0.00\% &   0.00\% & 100.00\% & 0.00\% &         61 & 152.11 \\
yo  &  98.46\% &  96.92\% & 0.00\% &  1.54\% &   1.54\% &   0.00\% & 0.00\% &         64 & 281.57 \\
gn  &  81.48\% &  81.48\% & 0.00\% &  0.00\% &   2.47\% &  16.05\% & 0.00\% &         81 & 234.95 \\
vec &  91.36\% &  91.36\% & 0.00\% &  0.00\% &   0.00\% &   8.64\% & 0.00\% &         81 & 184.90 \\
kw  &  91.57\% &  90.36\% & 0.00\% &  1.20\% &   3.61\% &   4.82\% & 0.00\% &         83 & 162.75 \\
\textbf{wuu} &   0.00\% &   0.00\% & 0.00\% &  0.00\% &  98.84\% &   1.16\% & 0.00\% &         86 & 157.15 \\
eml &  42.57\% &  42.57\% & 0.00\% &  0.00\% &   0.00\% &  57.43\% & 0.00\% &        104 & 177.88 \\
bh  &  89.42\% &  21.15\% & 0.00\% & 68.27\% &   1.92\% &   8.65\% & 0.00\% &        104 & 137.17 \\
min &  64.00\% &   6.00\% & 0.00\% & 58.00\% &  27.00\% &   9.00\% & 0.00\% &        180 & 649.85 \\
qu  & 100.00\% &  98.97\% & 0.00\% &  1.03\% &   0.00\% &   0.00\% & 0.00\% &        425 & 167.27 \\
su  &  99.00\% &  99.00\% & 0.00\% &  0.00\% &   0.00\% &   1.00\% & 0.00\% &        676 & 221.00 \\
jv  &  97.00\% &  86.00\% & 0.00\% & 11.00\% &   1.00\% &   2.00\% & 0.00\% &       2350 & 203.08 \\
als &  93.00\% &  93.00\% & 0.00\% &  0.00\% &   6.00\% &   1.00\% & 0.00\% &       7997 & 375.44 \\
la  &  98.00\% &  98.00\% & 0.00\% &  0.00\% &   2.00\% &   0.00\% & 0.00\% &      33838 & 224.11 \\
uz  &  98.00\% &  98.00\% & 0.00\% &  0.00\% &   2.00\% &   0.00\% & 0.00\% &      34244 & 369.99 \\
nds &  97.03\% &  95.05\% & 0.00\% &  1.98\% &   2.97\% &   0.00\% & 0.00\% &      35032 & 344.74 \\
sw  &  98.00\% &  98.00\% & 0.00\% &  0.00\% &   0.00\% &   2.00\% & 0.00\% &      40066 & 196.70 \\
br  & 100.00\% &  96.00\% & 0.00\% &  4.00\% &   0.00\% &   0.00\% & 0.00\% &      61941 & 239.56 \\
fy  &  97.00\% &  97.00\% & 0.00\% &  0.00\% &   2.00\% &   1.00\% & 0.00\% &      67762 & 340.23 \\
am  &  81.09\% &  79.10\% & 0.00\% &  1.99\% &  18.91\% &   0.00\% & 0.00\% &     287142 & 267.43 \\
af  & 100.00\% & 100.00\% & 0.00\% &  0.00\% &   0.00\% &   0.00\% & 0.00\% &     517353 & 339.18 \\
eu  & 100.00\% &  98.00\% & 0.00\% &  2.00\% &   0.00\% &   0.00\% & 0.00\% &    1099498 & 330.93 \\
mn  &  98.00\% &  94.00\% & 0.00\% &  4.00\% &   2.00\% &   0.00\% & 0.00\% &    1430527 & 309.94 \\
te  &  98.99\% &  93.94\% & 1.01\% &  4.04\% &   0.00\% &   1.01\% & 1.01\% &    1685185 & 412.31 \\
kk  & 100.00\% & 100.00\% & 0.00\% &  0.00\% &   0.00\% &   0.00\% & 0.00\% &    2719851 & 318.93 \\
ca  &  99.00\% &  91.00\% & 0.00\% &  8.00\% &   1.00\% &   0.00\% & 0.00\% &   13292843 & 333.38 \\
nl  &  98.00\% &  94.00\% & 2.00\% &  2.00\% &   2.00\% &   0.00\% & 4.00\% &  126067610 & 305.01 \\
it  &  87.13\% &  71.29\% & 1.98\% & 13.86\% &  11.88\% &   0.99\% & 1.98\% &  210348435 & 393.66 \\
zh  & 100.00\% &  97.00\% & 0.00\% &  3.00\% &   0.00\% &   0.00\% & 1.00\% &  232673578 & 195.60 \\
fr  & 100.00\% &  93.00\% & 0.00\% &  7.00\% &   0.00\% &   0.00\% & 5.00\% &  461349575 & 306.62 \\
es  & 100.00\% &  94.00\% & 0.00\% &  6.00\% &   0.00\% &   0.00\% & 3.00\% &  488616724 & 268.07 \\
en  &  99.00\% &  96.00\% & 0.00\% &  3.00\% &   0.00\% &   1.00\% & 1.00\% & 3809525119 & 364.65 \\
\bottomrule
\end{tabular}%
}
\caption{Audit results for a sample of 100 sentences from \textbf{OSCAR} for each language, compared to the number of sentences available in the dataset. If fewer than 100 sentences were available, all sentences were audited Language codes are as originally published. Length is measured in number of characters. Languages with less than 20\% correct sentences are boldfaced.}
\label{tab:oscar-full}
\end{table*}